\documentclass[acmsmall]{acmart}

\pdfoutput=1

\usepackage{todonotes}
\usepackage{subfigure}
\usepackage{calc}
\usepackage{amstext}
\usepackage{multicol}
\usepackage{multirow}
\usepackage{url}
\usepackage{caption}
\usepackage{makecell}
\usepackage{algorithm2e}
\usepackage{amsmath}

\SetCommentSty{mycommfont}

\DeclareMathOperator*{\argmin}{arg\,min}

\AtBeginDocument{%
  \providecommand\BibTeX{{%
    \normalfont B\kern-0.5em{\scshape i\kern-0.25em b}\kern-0.8em\TeX}}}

\begin{document}

% \title{A Survey on Semi-Supervised Learning for Delayed and Partially Labeled Data Streams}
% Dropped the "and" between Delayed and Partially, not to imply that we consider situations where only one of them happen. 
\title{A Survey on Semi-Supervised Learning for Delayed Partially Labelled Data Streams}

\author{Heitor Murilo Gomes}
\email{hgomes@waikato.ac.nz}
\affiliation{
  \institution{AI Institute, University of Waikato}
%   \country{New Zealand}
}

\author{Maciej Grzenda}
\email{m.grzenda@mini.pw.edu.pl}
\affiliation{%
  \institution{Faculty of Mathematics and Information Science, Warsaw University of Technology}
%   \country{Poland}
}

\author{Rodrigo Mello}
\email{mello@icmc.usp.br}
\affiliation{%
  \institution{ICMC, University of S\~{a}o Paulo}
%   \country{Brazil}
}

\author{Jesse Read}
\email{jesse.read@ polytechnique.edu}
\affiliation{%
  \institution{LIX, \'{E}cole Polytechnique, Institut Polytechnique de Paris}
}

\author{Minh Huong Le Nguyen}
\email{minh.lenguyen@telecom-paris.fr}
\affiliation{%
  \institution{T\'{e}l\'{e}com Paris, Institut Polytechnique de Paris}
}

\author{Albert Bifet}
\email{abifet@waikato.ac.nz}
\affiliation{%
  \institution{AI Institute, University of Waikato}
%   \country{New Zealand}
}


\begin{abstract}
Unlabelled data appear in many domains and are particularly relevant to streaming applications, where even though data is abundant, labelled data is rare. To address the learning problems associated with such data, one can ignore the unlabelled data and focus only on the labelled data (supervised learning); use the labelled data and attempt to leverage the unlabelled data (semi-supervised learning); or assume some labels will be available on request (active learning). 
The first approach is the simplest, yet the amount of labelled data available will limit the predictive performance. The second relies on finding and exploiting the underlying characteristics of the data distribution. The third depends on an external agent to provide the required labels in a timely fashion. This survey pays special attention to methods that leverage unlabelled data in a semi-supervised setting. We also discuss the delayed labelling issue, which impacts both fully supervised and semi-supervised methods. We propose a unified problem setting, discuss the learning guarantees and existing methods, explain the differences between related problem settings. Finally, we review the current benchmarking practices and propose adaptations to enhance them. 

\end{abstract}


\maketitle

\section{INTRODUCTION} \label{sec:introduction}

\begin{figure}[h]
% center everything in the figure
\centering
% horizontal node distance
\newcommand{\topdistA}{1cm}
\newcommand{\topdistB}{0.9cm}
\newcommand{\topdist}{1.5cm}
\resizebox{.55\columnwidth}{!}{
\begin{tikzpicture}[node distance=2cm]
% \title{Stream learning according to labels arrival time}
    \node(A1)                           {Data Stream Learning};
    \node(B1)       [below left=\topdistA and \topdistA of A1]      {Immediate};
    \node(B2)       [below =\topdistA and \topdistA of A1] {\textbf{Delayed}};
    \node(B3)       [below right=\topdistA and \topdistA of A1] {\begin{tabular}{c}
                                                                Never \\ \small{(Unsupervised)} \    \end{tabular}};
    \node(C1)       [below left=\topdistB of B2]       {\textbf{Fixed}};
    \node(C2)     [below right=\topdistB of B2]       {\textbf{Varying}};
    
    \node(D1)       [below =\topdistB of C1]       {\begin{tabular}{c}
                                                                All are labelled \\ \small{(Supervised)} \    \end{tabular}};
    \node(D2)       [below =\topdistB of C2]       {\begin{tabular}{c}
                                                                \textbf{Some are labelled} \\ \small{(Semi-Supervised)} \    \end{tabular}};
    \draw (A1) -- (B1);
    \draw (A1) -- (B2);
    \draw (A1) -- (B3);    
    \draw (B2) -- (C1);
    \draw (B2) -- (C2);
    \draw (B1) -- (D1);
    \draw (B1) -- (D2);    
    \draw (C1) -- (D1);
    \draw (C2) -- (D2);
    \draw (C1) -- (D2);
    \draw (C2) -- (D1);
\end{tikzpicture}
}
\caption{Learning from data streams according to labels arrival time, based on~\cite{ARF}. Highlighted in \textbf{bold} the dimensions associated with delayed partially labelled data streams.}
\label{fig:label_arrival}
\end{figure}
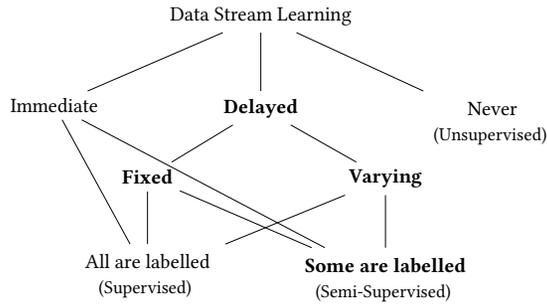

Situations where all the data are appropriately labelled, which allow us to perform supervised learning, are ideal, but many important problems are either unlabelled or only partially labelled. 
% Delayed verification
When dealing with streaming data, it is reasonable to expect some non-negligible verification latency, i.e. the label of an instance will be available sometime in the future, but not immediately. We identify data streams that exhibit both unlabelled data and verification latency as \textit{Delayed Partially Labelled Data Streams}. These characteristics refer to how (and if) labels are made available to the learning algorithm, as illustrated in Figure~\ref{fig:label_arrival}. 

A simple approach to cope with such data streams is to ignore both the unlabelled data and the labelling delay. Several methods were proposed, and evaluated, assuming a streaming scenario where all labels are immediately available~\citep{VFDT:2000,SEA, ONLINE_BAGGING_BOOSTING}. More recently, some authors investigated how to leverage unlabelled data using semi-supervised learning (SSL)~\citep{SS_CLUSTER_DS_HOSSEINI_2016,le2019semi}, or active learning~\citep{vzliobaite2013active}. On top of that, significant advances were made in modelling and analysing the impact of delayed labelling in supervised learning and concept drift detection for data streams~\citep{CD_DELAYED_ZLIOBAITE_2010,DELAYED_LABEL_TAXONOMY,grzenda2019delayed}. 

We focus the discussion on SSL methods for leveraging unlabelled data to enhance a supervised learning algorithm's predictive performance.  The basic assumption is that the algorithm has no influence over the labelling process, making active learning impractical. 
This work aims to organise the existing literature on SSL for data streams to facilitate new researchers and practitioners to navigate it. 
Concomitantly, we seek to elucidate the connections between the SSL and the delayed labelling literature to shed light on novel avenues for research. One challenging aspect of coping with delayed partially labelled data streams concerns the fair evaluation of algorithms. To assist in this perspective, we thoroughly discuss evaluation procedures for delayed partially labelled streams. This paper also aims to highlight the associations between related machine learning tasks, such as transductive learning, and to formalise the delayed partially labelled data streams. 

This survey extends the existing literature by focusing on SSL and delayed labelling for data streams. It is complementary to the vast literature on semi-supervised learning for stationary data~\citep{olivier2006ssl,SSLSurvey,van2020survey}; the evaluation of data streams~\citep{gama2013}, delayed labelling data streams~\citep{grzenda2019delayed} and SSL algorithms in general~\citep{REALISTIC_EVAL_SEMI_SUPERVISED_2018}; concept drift detection assuming immediate labelling~\citep{SURVEY_CD_GAMA_2014,CHAR_CD_WEBB_2016} or delayed labelling~\citep{CD_DELAYED_ZLIOBAITE_2010}; active learning for streaming data~\citep{SS_ACTIVE_DS_ZHU_2007,vzliobaite2013active}; and data stream mining in general~\citep{MINING_DS_REVIEW_2005,BOOK_MLDS_2017, gomes2019machine,bahri2021data}. 

The rest of this work is organised as follows. We first introduce the problem statement, clearly identifying similarities and differences with related problems in Section \ref{sec:problem_setting}. Next, in Section \ref{sec:learning_guarantees}, we point out theoretical learning guarantees for SSL in both offline and online scenarios. Section \ref{sec:methods} introduces existing SSL methods for streaming data. Section \ref{sec:comparative_analysis} includes a thorough discussion regarding the realistic assessment of SSL methods for data streams. The final Section \ref{sec:conclusion} concludes the paper and discusses avenues for future research as envisioned by the authors.

\section{PROBLEM DEFINITION} \label{sec:problem_setting}

In this section, we introduce the definitions and explicitly state assumptions concerning the problem setting. 
Precisely, we begin with a general definition of \textbf{supervised learning} and then describe \textbf{verification latency}, and \textbf{partially labelled data} in the context of \textbf{evolving data streams}. We devote the end of this section to discuss the related problems to the setting we introduce. 

% \todo[inline]{Shall we explicitly focus on classification in the abstract?}
% on the data stream classification task. 
\begin{definition} \label{def:X_sequence}
{\bf Instance data:} Let $X = \{x_{0}, \ldots, x_{1}, x_{\infty}\}$ represent an open-ended sequence of observations collected over time, containing input examples in which $x_k \in \mathbb{R}^n$ and $n \ge 1$.
\end{definition}

\begin{definition} \label{def:y_sequence}
{\bf Labels:} Let $y$ be an open-ended sequence of target values collected over time, such that for every entry in $y$ there is a corresponding entry in $X$, but the contrary may not be true, i.e., entries in $X$ without a corresponding entry in $y$ may exist.  Furthermore, when $y_k$ depicts a finite set of possible values, i.e., $y_k \in \{l_1, \ldots, l_L\}$ for $L \geq 2$, it is said to be a classification task, while when $y_k \in \mathbb{R}$ it denotes a regression task. 
\end{definition}
\begin{definition} \label{def:data_stream}
{\bf Data Stream:} Let $\Upsilon$ be a data stream i.e. a sequence of tuples $\mathcal{S}_1,\mathcal{S}_2,\ldots$ which includes two types of tuples i.e.
\[ \mathcal{S}_a =
  \begin{cases}
    \{(\mathbf{x}_k,?)\}       & \quad \text{if no true label is available yet}\\
   \{(\cdot,\mathit{y}_k)\} & \quad \text{when a true label for $\mathbf{x}_k$ becomes available}\\
  \end{cases}
\]
Hence, $\{(\mathbf{x}_k,?)\}$ is a tuple containing the observation, whereas $\{(\cdot,\mathit{y}_k)\}$ is a tuple containing the label corresponding to this observation.
\end{definition}

% \begin{definition} \label{def:instance_indicator}
% {\bf Instance indicator function:} Let $I(\mathcal{S}_a)$ denote instance  indicator function. 
% \[ I(\mathcal{S}_a) =
%   \begin{cases}
%           1 & \quad \mathcal{S}_a=\{(\mathbf{x}_k,?)\} \\
%   0 & \quad \mathcal{S}_a=\{(\cdot,\mathit{y}_k)\}\\
%   \end{cases}
% \]
% \end{definition}

% \begin{definition} \label{def:label_indicator}
% {\bf Label indicator function:} Let $L(\mathcal{S}_a)$ denote label  indicator function, defined as $L(\mathcal{S}_a)=1-I(\mathcal{S}_a)$.
% \end{definition}

\begin{definition} \label{def:time_function}
{\bf Temporal-Mapping function:} Let $T(\cdot)$ denote a function that extracts the precise discrete time unit $t$ that $x_k$ and $y_k$ became available. It is relevant to mention that $T(x_k) \leq T(y_k)$ must always hold, indicating that the input data $x_k$ becomes available at same the moment or before $y_k$. 
\end{definition}

\begin{definition} \label{def:stream_section}
{\bf Stream section:}
Let $\Psi[T_\mathrm{min},T_\mathrm{max}]$ denote a stream section i.e. a sequence of instances and true labels  that became available during a time window $[T_\mathrm{min},T_\mathrm{max}]$. This means, $\forall x_k,y_k\in \Psi[T_\mathrm{min},T_\mathrm{max}]: 
(T_\mathrm{min}\leq T(x_k)\leq T_\mathrm{max}) \land (T_\mathrm{min}\leq T(y_k)\leq T_\mathrm{max})$.
\end{definition}

\begin{definition} \label{def:verification_latency}
{\bf Verification latency:} Let $V(x_k, y_k)=T(y_k) - T(x_k)$ represent the time difference a.k.a. ``verification latency'' of the labelled instance represented by the tuple $(x_k, y_k)$. 
\end{definition}

\begin{definition} \label{def:inf_verification_latency}
{\bf Infinitely delayed labels:} Let $V(x_k, y_k)=\infty$ denote the verification latency of an infinitely delayed labelled instance a.k.a. unlabelled instance. 
\end{definition}

$T(x_k) = T(y_k)$, as seen in Definition~\ref{def:time_function}, denotes a situation where both the input example and its label are provided at the same time instant, what is the same as receiving training instances from some batch learning task. Asymptotically, $V(x_k, y_k) \to \infty$ so that an observation $x_k$ has no corresponding label $y_k$ (see Definition~\ref{def:verification_latency}). 

Based on the aforementioned definitions:
\begin{itemize}
    \item \textbf{(i) Immediate and fully labelled.} $\forall x \in X \land \forall y \in Y$ $T(y) - T(x) = 1$, i.e., the verification latency between $x$ and $y$ corresponds exactly one time unit. 
    \item \textbf{(ii) Delayed and fully labelled.} $\forall x \in X \land \forall y \in Y$ $T(y) - T(x) = D$, where $D$ is a random variable representing the discrete delay between $x$ and $y$ limited by the finite range $D \in \mathcal{Z}_+$, where $\max(D)$ denotes the maximum delay. 
    % \item \textbf{(iii) Delayed weakly labelled.} \todo[inline]{Maybe this one could be added to the previous definition? Maciej: I think the key question is whether to use {\em weekly labelled} also to refer to single label (rather than multi-label) setting}
    \item \textbf{(iii) Immediate and partially labelled.} If we relax the constraint that every $X$ has a corresponding entry on $Y$, we obtain a setting where $X$ is only partially labelled. It is useful to emphasize the difference between entries in $X$ which will be labelled as $X_L$ and those that will not be labelled as $X_U$, and also to ascertain that often $|X_L| \ll |X_U|$ as the labelling process can be costly. 
    \item \textbf{(iv) Delayed and partially labelled.} Similarly to (iii), we extend (ii) such that labels are delayed and some of them never arrived, i.e. they are infinitely delayed. 
    
    % \todo[inline]{Maciej: I think adding this option will be a way to explicitly refer to what we promise in the title.}
\end{itemize}

The majority of the literature with respect to semi-supervised learning for data streams has been devoted to (iii), while the intersection between delayed and partially labelled data, as in (iv), is yet to be thoroughly explored. 
Besides the matters of label availability, another concept that is worth discussing in our definitions is whether the data distribution is \textbf{stationary} or \textbf{evolving}. In general, we assume evolving data distributions, thus \textbf{concept drifts} are deemed to occur, which may inadvertently influence the decision boundaries, and affect learned models. 
Note that if a concept drift is accurately detected (without false negatives) and dealt with by fully or partially resetting models as appropriate an independent and identically distributed (iid) assumption can be made (on a per-concept basis), since each concept can be treated as a separate iid stream, thus a series of iid streams to be dealt with\footnote{Not in every case a concept drifting stream can be decomposed into a sequence of iid streams. Theoretically, gradual (or incremental) drifts may occur where the distribution changes after every instance.}. Nevertheless, the typical nature of a data-stream as being fast, dynamic and partially labelled encourages the in-depth study of methods for properly evaluate algorithms under these settings and semi-supervised algorithms to exploit unlabelled data. 
% \todo[inline]{Maybe we can add here that not in every case a concept drifting stream can be decomposed into a sequence of iid streams, since gradual drifts may occur. Theoretically the distribution could change with every instance?}

% \subsection{Challenges} \label{sec:challenges}
% We now present some practical challenges when addressing delayed and partially labelled data streams with the intent of improving supervised learning algorithms. 

% \begin{itemize}
%     \item if the label changes in the meantime?
%     \item If the concept drifts, should the label still be used for training or testing?
%     \item \textbf{Technical} challenges
%     \begin{itemize}
%         \item store input instances for how long until label arrives? 
%         \item should we use a y that took too long to arrive? 
%         \item how to match y to their corresponding X (an unique id?)
%         \item event time X processing time. 
%         \item Out of order data. 
%         \item whether to use evolving predictions
%     \end{itemize}
%     % \item Theoretical challenges 
%     % \begin{itemize}
%     %     \item non-iid data
%     %     \item limitations of unlabelled data for leveraging learning models
%     % \end{itemize}
%     % \item unsupervised drift detection (when labels arrive, how to use them to improve detection?)
% \end{itemize}

\subsection{Related Problems} \label{sec:related_problems}
% Moved the active learning description to here
In this section, we provide a short description of learning problems that are related to SSL for data streams, but that are not further scrutinized in this paper to avoid diverging from the delayed partially labelled problem setting. 

\textbf{Active learning.} When dealing with an abundant amount of missing labels or a costly labelling procedure, \textit{active learning} can be a viable alternative to SSL. Active learning approaches attempt to overcome the labelling bottleneck by querying the label of selected unlabelled instances to an oracle, such that the instances to be labelled are the most uncertain (e.g. a point lying close to the discriminative hyperplane) and that the answered labels can bring the highest value to the learning process. In this way, the active learner aims to achieve high accuracy using as few labelled instances as possible, thereby minimizing the cost of collecting labelled data~\citep{ActiveLearning}. 
\citet{vzliobaite2013active} introduced a theoretical framework for active learning from drifting data streams. The authors stated three conditions for a successful active learning strategy in a drifting scenario: balancing the labelling budget over an infinite amount of time, perceiving changes anywhere in the instance space, and preserving the distribution of incoming data for detecting changes. Furthermore, in \citep{vzliobaite2013active} three strategies were presented and empirically evaluated, assuming that an external adaptive learning algorithm is present. 

Despite the advances in active learning for streaming data, it is sometimes hard to employ such strategies. The first reason is that the oracle's response time may be too slow, as it often relies on a human expert. Second, still related to the labelling response time, if a concept drift occurs, the instances selected to be labelled may be outdated. The latter issue can be amended by using active learning strategies that take drift into account, as shown in \citep{vzliobaite2013active}. 
Besides the issues involving the instability of the concepts, and delay to obtain the labels, \citet{SS_ACTIVE_DS_ZHU_2007} also discusses the challenges related to the pool of candidates (instances to be labelled) being dynamic and issues related to the volume of data. To address these challenges, \citet{SS_ACTIVE_DS_ZHU_2007} proposed an ensemble-based active learning classifier, where the minimization of the ensemble variance guides the labelling process.

\textbf{Transductive learning.} Transductive learning concerns a situation where the unlabelled test data set contains the whole of instances to be predicted, thus instead of producing a general model for predicting any future instance, the output is the predictions. This is a ``closed world'' assumption, where a successful solution is one where the algorithm can approximate the true labels of the instances solely for the finite test data set. This differs from inductive learning, where the goal is to yield a learning model capable of generalizing to previously unseen instances. 
Transduction is a powerful technique to leverage unlabelled data, but it is limited to situations where the goal is to produce accurate predictions to a given set of instances and not devise a general rule. The majority of the algorithms for stream learning tend to focus on inductive learning. One possible reason is that traditional transductive methods require many computations. Thus frequently performing these may be prohibitive in a stream setting where predictions are often required to be fast. To circumvent this problem, \citet{ho2004learning} proposed an incremental version of the transductive confidence machine (TCM)~\citep{gammerman1998learning}. 
However, even though it is feasible to alleviate the computational aspects, another essential issue is that since data streams are unbounded, it is challenging to generate a closed set of instances. 
%% Mention this again in the Discussion. New methods for transductive learning for streaming data? 

\textbf{Weakly multi-labelled data}.  
Semi-supervised learning often stems from the case of having limited human labelling power to label all examples. Such a scenario is particularly inherent to data streams, where there are many instances, and they are arriving continuously. It is also aggravated when there are multiple label variables associated with each input -- the so-called \emph{multi-label} learning problem \citep{Overview}. In this case, multiple labels $y_k \subseteq \{l_1,\ldots,l_L\}$ (i.e., a \emph{subset}) are associated with each instance. In this context, \emph{weakly labelled} data (see, e.g., \citep{WeakLabels}) refers to instances where some, but not all of the relevant/true labels, have been applied to an instance. Specifically, the absence of a label in this subset does not necessarily imply that it is not relevant; and this is the challenge: to identify which of the non-relevant labels are missing in the labelled examples (it is not clear which ones are missing). A related concept of partial multi-label learning \citep{PartialLabels} is the generalization that additionally accounts for the possibility of false-positives (labels signalled as relevant, which are actually not). If we view a subset as binary indicator variables (as is typical in the literature), these problems become equivalently to $L$ parallel (and possibly interdependent) noisily-labelled streams. Similar issues exist in the general multi-output case (extending to regression) \citep{UnifiedView,PRC}.

\textbf{Missing values}. 
Weakly multi-labelled data is also related to having \emph{missing values} in the output/label space, except in this latter case it is clear which values are missing. This can be illustrated with an example in vector notation: $y_k = \{l_1,l_3\} \Leftrightarrow [1,0,1,0]$ (supposing $L=4$) where in the missing-valued case we may have $[?,0,1,?]$ (compared to the weakly-labelled case where, e.g., $[0,0,1,0]$ where $l_1$ is a false negative in our label set). Of course, it is also common to have missing values in the input space (as this affects all kinds of machine learning). This context is not a main focus of this survey. However, we note that a common method to deal with missing values is imputation. And, by building classifier or regression models to carry out this imputation (according to the variable domain being imputed), it is possible to frame a missing value imputation as a weakly-labelled multi-label problem \citep{Cascade}; which in turn can be seen as $L$ parallel partially-labelled streams. %This is very much related to the problem of having noisy values. 

\textbf{Initially Labelled Streaming Environment} 
Labelled data may only be available at the beginning of the learning process. Therefore, a supervised learning algorithm can be trained with the initial data, and another unsupervised mechanism used to update the model during execution. This is a challenging problem setting as new labelled data is not available throughout execution, therefore it is not possible to confidently verify the accuracy of the model during execution. This setting was explored by \citet{krempl2011algorithm}, where the APT algorithm was proposed to track concept drifts even in the absence of labelled data. Later, \citet{dyer2014} proposed the COMPOSE framework to tackle the same problem setting, which also featured a detection mechanism that was independent of labelled data. 

\textbf{Few-shot learning}. 
Few-shot learning \citep{FewShotLearning} refers to feeding the learning algorithm with very few amount of labelled data. This approach is very popular in fields such as computer vision, where a model for object categorization is able to discern between different objects even without having multiple instances of each object for training. 
The term few-shot is accompanied by low-shot, 1-shot and 0-shot, which refer to training with a low amount of instances per class, only one per class and not even one labelled instance for each class, respectively. As expected, as the number of labelled instances shrinks, the harder to produce accurate models. Approaching few-shot learning (and its variants) using semi-supervised learning is a common technique, also, when possible it is usual to leverage pre-trained models from similar domains (transfer learning). 

% Few-shot learning refers to training a learning model with a scarce amount of labelled data for a given supervised learning task. One approach to deal with few-shot learning problems is to exploit data beyond the available labelled data. The problem then becomes on how to obtain this extra data. Depending on the domain, one can use an external data source to augment the dataset, thus effectively performing transfer learning. This is a very popular approach in domains related to computer vision tasks. Another approach is to leverage the unlabelled data in a semi-supervised manner.
% Finally, one can also augment the dataset by transforming the labelled data, for example, by adding random noise to it and generating new samples. 

\textbf{Concept evolution}. 
In some problems, the number of labels may vary over time. This problem is known as concept evolution~\citep{CONCEPT_EVOLUTION_MASUD_2010}. Concept evolution characterizes a challenging problem where some instances are not only unlabelled but belong to a class that has not yet been identified. This is true for scenarios where one want to characterize malware per family instead of the comparatively more manageable task of classifying applications into malware or ``goodware'' (binary classification). 
In this survey, we do not approach such a problem as it requires a different definition of the problem as not all class labels are known \textit{a priori}. A practical approach to address concept evolution in data streams is to leverage the clustering assumption~\citep{van2020survey} or apply novelty (i.e. anomaly) detection techniques to identify novel classes. \citet{masud2010classification} introduced DXMiner, an algorithm capable of detecting novel classes by identifying novel clusters, while \citet{masud2010addressing} used an outlier detector and a probabilistic approach to detect novel classes. \citet{abdallah2016anynovel} proposes a method to continuously monitor the flow of the streaming data to detect concept evolutions, whether they are normal or abnormal. 

% such as clustering new unlabelled instances and identifying new clusters as novel classes, this approach was used in 

% \subsection{DRAFT: Assumptions}
% \begin{itemize}
%     \item smoothness
%     \item low-density
%     \item manifold
% \end{itemize}

\section{LEARNING GUARANTEES} \label{sec:learning_guarantees}

% Theoretical aspects
Supervised learning relies on different theoretical frameworks to ensure the conditions under which learning is guaranteed, being the Statistical Learning Theory (SLT) the most prominent contribution~\citep{vapnik2013nature}. According to SLT, supervised learning is defined as the process involved in converging to the best as possible classification or regression function $f:\mathcal{X} \to \mathcal{Y}$ contained in the algorithm bias $\mathcal{F}$, a.k.a. its space of admissible functions, in which $X$ corresponds to the input space and $Y$ to the output space containing labels.

This convergence process is essentially focused on approaching some loss measurement $R_\text{emp}(f)$ (or empirical risk) computed on training examples $(x_i,y_i) \in X \times Y$ to its expected value $R(f)$ (or risk) which is only computable by having the joint probability distribution (JPD) $P(X, Y)$. The basic and most important concept behind this convergence is to make possible the use of the empirical risk $R_\text{emp}(f)$ as a good estimation for the risk $R(f)$, provided $f \in \mathcal{F}$. Observe that by making sure $R_\text{emp}(f) \to R(f)$ and the training sample size $n \to \infty$, one can use the empirical risk to select the best classification function $f^*$ by using:
\begin{align*}
    f^* = \argmin_{f \in \mathcal{F}} R_\text{emp}(f),
\end{align*}
assuming the impossibility of computing the risk $R(f)$ for real-world problems, because we would never have access to the JPD.

Based on the Law of Large Numbers~\citep{books/sp/DevroyeGL96}, Vapnik~\citep{vapnik2013nature} formulated the Empirical Risk Minimization Principle (ERMP) to represent $R_\text{emp}(f) \to R(f)$ as $n \to \infty$ in form:
\begin{align}
    P(\sup_{f \in \mathcal{F}} |R_\text{emp}(f) - R(f)| > \epsilon) \leq 2 \mathcal{N}(\mathcal{F},2n) \exp{(-2 n \epsilon^2)},
\end{align}
given $f$ is selected from the algorithm bias $\mathcal{F}$, the supremum reinforces the worst possible classifier that most influences in the divergence between both risks, $\mathcal{N}(\mathcal{F},2n)$ is the shattering coefficient or growth function defining the number of distinct classifications built from $\mathcal{F}$, $R_\text{emp}(f), R(f) \in [0,1]$ and $\epsilon \in \mathbb{R}_+$.

Given the use of the Law of Large Numbers, a set of assumptions must be ensured to prove learning, otherwise the ERMP becomes inconsistent. The first assumption is that the JPD $P(X, Y)$ is fixed, so it does not change along with the data sampling, otherwise the convergence could not be ensured given samples would follow a different probability distribution. Second, all samplings obtained from $P(X, Y)$ must be independent of one another and identically distributed so that every possible event from JPD will have its probability of being chosen as defined by its corresponding density.

It is relevant to mention that SLT can be mapped into other theoretical frameworks such as PAC-Learning and regularization methods~\citep{series/hhl/LuxburgS11}. Thus, from a such theoretical point of view, the following sections assess learning guarantees for both semi-supervised offline and online scenarios.

\subsection{Semi-supervised learning in offline scenarios}

From the perspective of the semi-supervised learning on offline scenarios, the assumptions after the Law of Large Numbers can be still met depending on the target application, to mention: (i) the joint probability distribution (JPD) $P(X, Y)$ must be fixed, and (ii) samplings from such JPD must be independent from each other. From such theory, if the JPD changes over time, we could somehow manage to obtain as much guarantee as possible so that the Empirical Risk Minimization Principle (ERMP) becomes partially consistent, and thus we can come up with some learning bounds. Complementary, if instances are not independent from one another, one option is to restructure data spaces as discussed in~\citep{DBLP:journals/eswa/PagliosaM17}.

% Having said that, in this section we consider that our semi-supervised offline scenario is represented by a single, and thus fixed, JPD whose data instances are independent from each other, while next section takes into account the opposite scenario that is commonly found in online learning. 
% Suggestion
In this section, we consider that our semi-supervised offline scenario is represented by a single, and thus fixed, JPD whose data instances are independent of each other, while the next section considers the opposite scenario common in online learning.
Therefore, let us have some dataset $(x_i,y_i) \in X \times Y$, for $i = 1, \ldots, n$, containing $n$ input examples $x_i$ and their corresponding class labels $y_i = \{\empty,-1,+1\}$ with three possibilities: a negative, a positive and an empty label information. Consider $\empty$ as the absence of a class label so that one has no information about such instance, consequently its relative misclassification cannot be computed using a loss function $\ell(x_i, y_i, f(x_i))$ provided a classifier $f$. The absence of class labels is what makes this scenario be defined as a semi-supervised learning task, otherwise it would be a typical supervised task.

If we had all class labels, so that $y_i = \{-1,+1\}$, the ERMP after the SLT would be sufficient to represent learning bounds which formulates the conditions for which the empirical risk approaches the expected risk, i.e., $R_\text{emp}(f) \to R(f)$, as the sample size $n \to \infty$:
\begin{align}\label{ineq:ermp}
    P(\sup_{f \in \mathcal{F}} |R_\text{emp}(f) - R(f)| > \epsilon) \leq 2 \mathcal{N}(\mathcal{F},2n) \exp{(-2 n \epsilon^2)},
\end{align}
given the empirical risk is computed on a sample:
\begin{align*}
    R_\text{emp}(f) = \frac{1}{n} \sum_{i=1}^{n} \ell(x_i, y_i, f(x_i)),
\end{align*}
and the expected risk on the JPD:
\begin{align*}
    R(f) = \mathbb{E}(\ell(X, Y, f(X))).
\end{align*}

In the same supervised scenario, Vapnik~\cite{vapnik2013nature} proved the Generalization bound from Inequation~\ref{ineq:ermp} as follows:
\begin{align}\label{ineq:gb}
    R(f) \leq R_\text{emp}(f) + \sqrt{\frac{4}{n} \left( \log(2 \mathcal{N}(\mathcal{F},2n)) - \log{\delta} \right)},
\end{align}
for $\delta = 2 \mathcal{N}(\mathcal{F},2n) \exp{(-2 n \epsilon^2)}$, so that one can estimate how far the expected risk is from the risk computed on some sample plus a divergence defined by the squared-root term.

Once we do not have access to a fully labeled dataset so that $y_i = \{\empty,-1,+1\}$, we relax the learning bounds provided by the ERMP by redefining the sample size $n = \sum_{i}{\mathbf{I}(y_i)}$ as the number of labeled instances according to the indicator function:
\begin{align*}
    \mathbf{I}(y_i):={
            \begin{cases}
                1&{\text{if }}y_i \in \{-1,+1\},\\
                0&{\text{otherwise}}.
            \end{cases}
            },
\end{align*}
considering all available examples in some dataset $(x_i,y_i) \in X \times Y$, for $i = 1, \ldots, n_\text{all}$, from which the same SLT bounds (see Inequations~\ref{ineq:ermp} and~\ref{ineq:gb}) can be ensured provided the JPD $P(X, Y)$ is fixed and data instances are iid.

In attempt to show how this first and straight-forward conclusion is useful to define learning bounds for semi-supervised learning in offline scenarios, consider some dataset $(x_i,y_i) \in X \times Y$, for $i = 1, \ldots, n_\text{all}$ instances sampled from $P(X, Y)$. Consider that a fraction $\nu$ of instances was pre-labeled in $\{-1,+1\}$ so that $n = \nu \times n_\text{all}$, therefore considering some pre-defined $\delta$ as the upper probability bound for (see Inequation~\ref{ineq:ermp}):
\begin{align}
    P(\sup_{f \in \mathcal{F}} |R_\text{emp}(f) - R(f)| > \epsilon) \leq \delta,
\end{align}
we can study the minimal training sample size~\cite{mello2019shattering} to ensure such bound which relies on $\delta = 2 \mathcal{N}(\mathcal{F},2n) \exp{(-2 n \epsilon^2)}$ as proved by Vapnik~\cite{vapnik2013nature}. In that sense, let us assume the Shattering coefficient function $\mathcal{N}(\mathcal{F},2n) = n^2$ for a specific semi-supervised algorithm working on some $d$-dimensional Hilbert input space, thus defining the maximal number of distinct classifications as the sample size $n$ grows. Given the shattering coefficient represents the complexity of the algorithm bias, term $\delta$ reflects such complexity in terms of the pre-labeled sample size $n$ available for computing the loss function $\ell(x_i, y_i, f(x_i))$ provided every classifier $f \in \mathcal{F}$.

Thus, after assuming $\mathcal{N}(\mathcal{F},2n) = n^2$, we compute $\delta$ as follows:
\begin{align}
    \delta  = 2 \mathcal{N}(\mathcal{F},2n) \exp{(-2 n \epsilon^2)} = 2 n^2 \exp{(-2 n \epsilon^2)} = 2 \exp{(2 \log{n} -2 n \epsilon^2)},
\end{align}
from which we can analyze the minimal labeled sample size $n$, characterizing the acceptable divergence between the empirical risk $R_\text{emp}(f)$ and its expected value $R(f)$. As this scenario considers a polynomial shattering coefficient, this curve produced by $\delta$ as $n$ varies will approach zero. For instance, if we decide to accept a divergence of $5\%$ ($\epsilon = 0.05$) and set $\delta=0.1$ to have a probability of getting both risks acceptably close with probability of $0.9$ ($90\%$ of the cases), we would need $n=3,908$ training labeled examples. Consequently, one can find the minimal training sample size to ensure a given divergence $\epsilon$ and some probabilistic upper bound $\delta$. For the sake of comparisons, we suggest to consider at least $\epsilon \leq 0.05$ and $\delta \leq 0.05$.

It is still worth to discuss how an algorithm bias changes the minimal training sample size necessary to address some semi-supervised task. The more complex an algorithm is, the steeper will be its shattering coefficient curve thus directly requiring more data instances to ensure the same learning guarantees, provided $\epsilon$ and $\delta$. As discussed in~\cite{mello2019shattering}, such complexity is related to the number of hyperplanes used to devise a proper decision boundary, the number of dimensions the input space has, as well as other factors such as how the dataset is organized (e.g. graph or table of variables). The need for estimating the shattering coefficient function to proceed with further algorithmic analysis has been motivating several studies in the last years~\cite{mello2019measuring,mello2019shattering}~\footnote{We suggest the following R Package to estimate the Shattering coefficient function -- \url{https://cran.r-project.org/package=shattering}.}.

Alternatively, we may consider the Generalization bound (see Inequation~\ref{ineq:gb}) to study a model we wish to induce from data. In that circumstance, assume we have estimated the shattering coefficient $\mathcal{N}(\mathcal{F},2n) = n^2$ after setting $\delta=0.05$, then we have:
\begin{align*}
    R(f) &\leq R_\text{emp}(f) + \sqrt{\frac{4}{n} \left( \log(2 \mathcal{N}(\mathcal{F},2n)) - \log{\delta} \right)} = R_\text{emp}(f) + \sqrt{\frac{4}{n} \left( \log(2 n^2) - \log{0.05} \right)},\\
    R(f) &\leq R_\text{emp}(f) + \sqrt{\frac{8 \log(n)}{n} + \frac{14.7555}{n}},
\end{align*}
from which we conclude the empirical risk $R_\text{emp}(f)$ diverges from its expected value $R(f)$ according to term $\sqrt{\frac{8 \log(n)}{n} + \frac{14.7555}{n}}$, which naturally converges to:
\begin{align*}
        \lim_{n \to \infty} \sqrt{\frac{8 \log(n)}{n} + \frac{14.7555}{n}} = 0,
\end{align*}
as the training sample size $n$ tends to infinity, thus proving learning in such a semi-supervised scenario. However, we may consider what such square-root term brings in terms of information and comparison among different learning settings. 
%Of course, the greater such term is, the more complex is the algorithm bias and the necessary training sample size to ensure learning bounds.
% Suggestion
Consequently, the greater such term is, the more complex the algorithm bias and the necessary training sample size to ensure learning bounds.

The square-root term represents the variance provided the space of admissible functions $\mathcal{F}$. It consequently relates to the cardinality of the classification functions enclosed in the algorithm bias and the acceptable upper bound for the ERMP (see Inequation~\ref{ineq:ermp}) given $\delta = \mathcal{N}(\mathcal{F},2n) \exp{(-2 n \epsilon^2)}$, being therefore a way of regularizing the learning process. Regularization strategies are used to reduce the error by fitting an appropriate set of functions given some training set, consequently avoiding overfitting~\cite{series/hhl/LuxburgS11}.

\subsection{Semi-supervised learning in online scenarios}

% Artigos:
% Does Unlabeled Data Provably Help? Worst-case Analysis of the Sample Complexity of Semi-Supervised Learning \cite{balcan200621}
% An Augmented PAC Model for Semi-Supervised Learning \cite{ben2008does}
% A survey on semi-supervised learning \cite{van2020survey}
From the perspective of the semi-supervised learning on online scenarios, the assumptions after the Law of Large Numbers (LLN) must be somehow dealt with, to mention once more: (i) the joint probability distribution (JPD) $P(X, Y)$ must be fixed, and (ii) samplings from such JPD must be independent from one another. We easily conclude that both assumptions limit online learning in which we certainly expect the JPD to change over time, a classical aspect known as concept drift in the data streams scenario, as well as data observations will most certainly present some degree of dependence. Therefore, some strategy must be employed to still make the Empirical Risk Minimization Principle (ERMP) consistent so learning can be theoretically ensured.

As proposed by Pagliosa and Mello~\cite{DBLP:journals/eswa/PagliosaM17}, Dynamical system tools can be used to reconstruct the input space $X$ so that all dependences are represented in terms of a new set of dimensions. They employ Takens' embedding theorem~\cite{10.1007/BFb0091924} to reconstruct some unidimensional time series $S = \{s_1, \ldots, s_t\}$ into some high dimensional space referred to as phase space $\Phi$ whose points or states $\phi_t \in \Phi$ are in form:
\begin{align*}
    \phi_t = (s_t, s_{t+\tau}, s_{t+2\tau}, \ldots, s_{t+(m-1) \times \tau}),
\end{align*}
given $\tau$ refers to the necessary time delay to unfold the temporal relationships, a.k.a. time delay, and $m$ corresponds to the embedding dimension or simply the number of axes necessary to represent all dependencies.

% According to their approach, a single-dimensional data stream could be reconstructed into some phase space so that their temporal dependencies would be represented therefore all states $\phi_t \in \Phi$ would be independent from one another, thus solving Assumption (ii) of the LLN if one needs to perform some regression on unidimensional data, however it leaves some important open questions: (a) how to deal with multidimensional data streams?; and (b) how to deal with the classification task of semi-supervised data streams?
% Suggestion
According to their approach, a single-dimensional data stream could be reconstructed into some phase space so that their temporal dependencies would be represented; therefore, all states $\phi_t \in \Phi$ would be independent of one another, thus solving Assumption (ii) of the LLN if one needs to perform some regression on unidimensional data. However, it leaves some important open questions: (a) how to deal with multidimensional data streams?; and (b) how to deal with the classification task of semi-supervised data streams?

% Question (a)
Question (a) associated with Assumption (ii) was previously answered by Serra et al.~\cite{Serr__2009} who used the same concepts from dynamical systems to reconstruct multivariate time series $S$ for $s_t \in \mathbb{R}^d$, given $d > 1$, as follows (the upper index corresponds to each variable composing the multivariate time series $S$):
\begin{align*}
    \phi_t = (s_t^1, s_{t+\tau}^1, \ldots, s_{t+(m-1) \times \tau}^1, s_t^2, s_{t+\tau}^2, \ldots, s_{t+(m-1) \times \tau}^2, \ldots, s_t^d, s_{t+\tau}^d, \ldots, s_{t+(m-1) \times \tau}^d),
\end{align*}
so that, in addition to represent the temporal relationships of a single variable with itself, it also unfolds the dependencies that each variable of the time series (upper index) has with the others. Therefore, assuming that a multidimensional data stream has some time index as data observations arrive, one can extend Serra's framework to solve Assumption (ii) of the LLN, thus answering the first question. Observe it is not an unreasonable assumption to require data observations are indexed over time.

% Fenômenos com data dependence
Observe that someone may doubt the presence of data dependence among stream observations. However, it is not difficult to mention several real-world phenomena illustrating such scenarios, such as in the context of air temperatures of a given world region, climatic variables, interaction of chemicals in reactions and the growth of populations along time~\cite{kantzbook}. Several researchers associated with the area of dynamical systems have been applying the same tools to obtain iid spaces~\cite{DBLP:conf/bracis/ReadRNM20,Karimov_2020,Ushio2020}.

Now, we get back to the Assumption (i) of the LLN, which requires the joint probability distribution (JPD) $P(X,Y)$ to be fixed in order to ensure learning. In that specific scenario, we suggest modelling the current JPD using past data observations and, as soon as some relevant data distribution change or drift is identified, past data must be discarded and the learning algorithm must start buffering new observations to induce a new model. Such approach is used in~\cite{DBLP:journals/eswa/MelloVFB19} to build up new models as data arrives, using McDiarmid's inequality to detect data drifts and indicate the best moment to retrain learning models using recent collected observations.

At last, all theoretical concepts addressed in this section intend to support other researchers to analyze their learning algorithms in an attempt of obtaining as much as possible guarantees while tackling partially labelled real-world streaming problems.

\section{METHODS} \label{sec:methods}

Supervised machine learning is defined by using labelled data to train algorithms to predict unseen and unlabelled instances. These unlabelled examples do not influence algorithm anyhow.
In most applications, obtaining labelled data is time-consuming and expensive, as labelling often depends upon human annotators. On the other hand, acquiring unlabelled data is an easier task, but these data cannot update supervised models directly. 
Semi-supervised learning is a paradigm of learning that exploit unlabelled data to leverage models trained with labelled data. 
%, to leverage supervised models relying solely on a small amount of labelled data and a large volume of unlabelled data. 

\begin{figure}
	\includegraphics[width=0.49\textwidth]{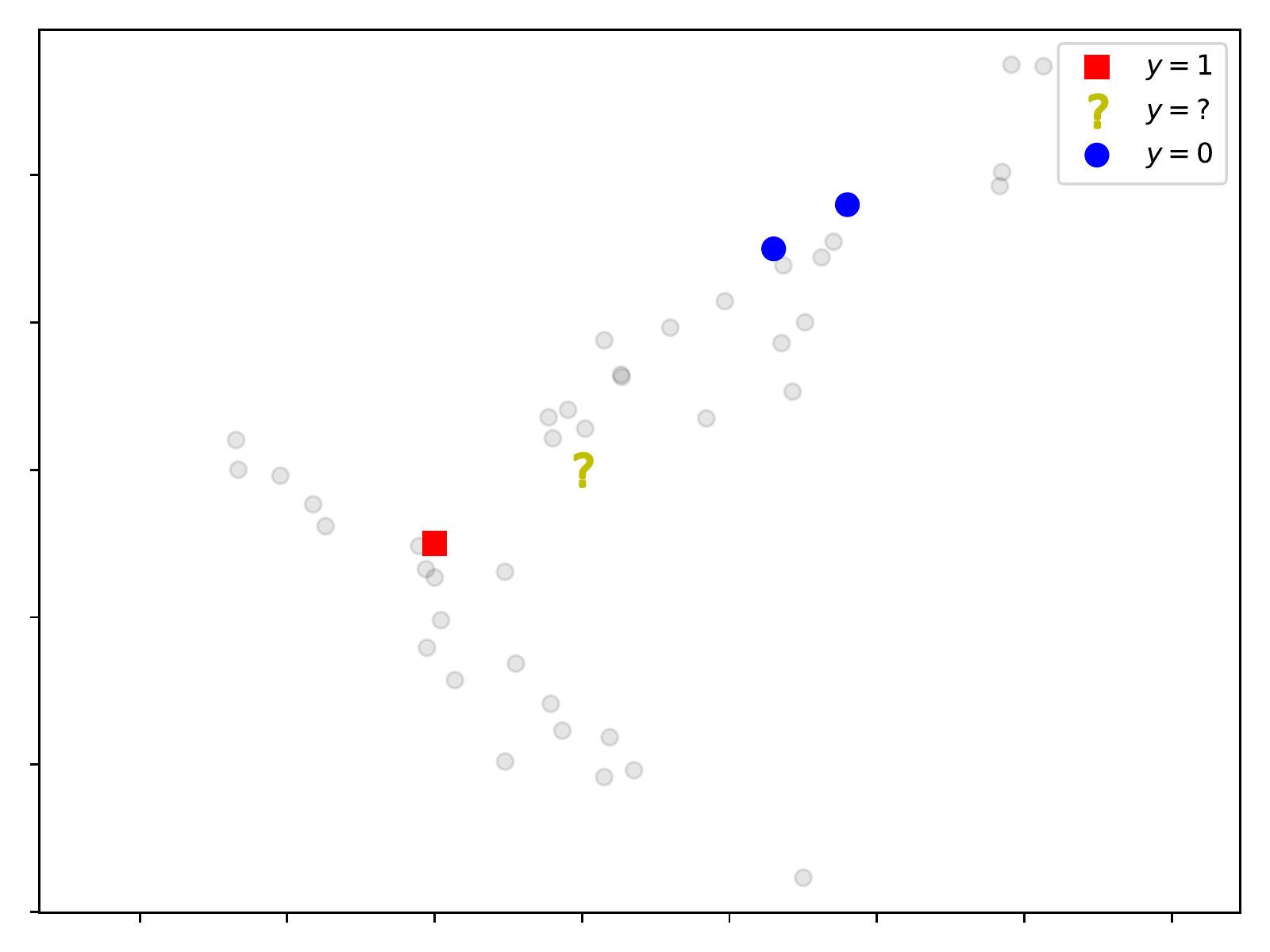}
	\includegraphics[width=0.49\textwidth]{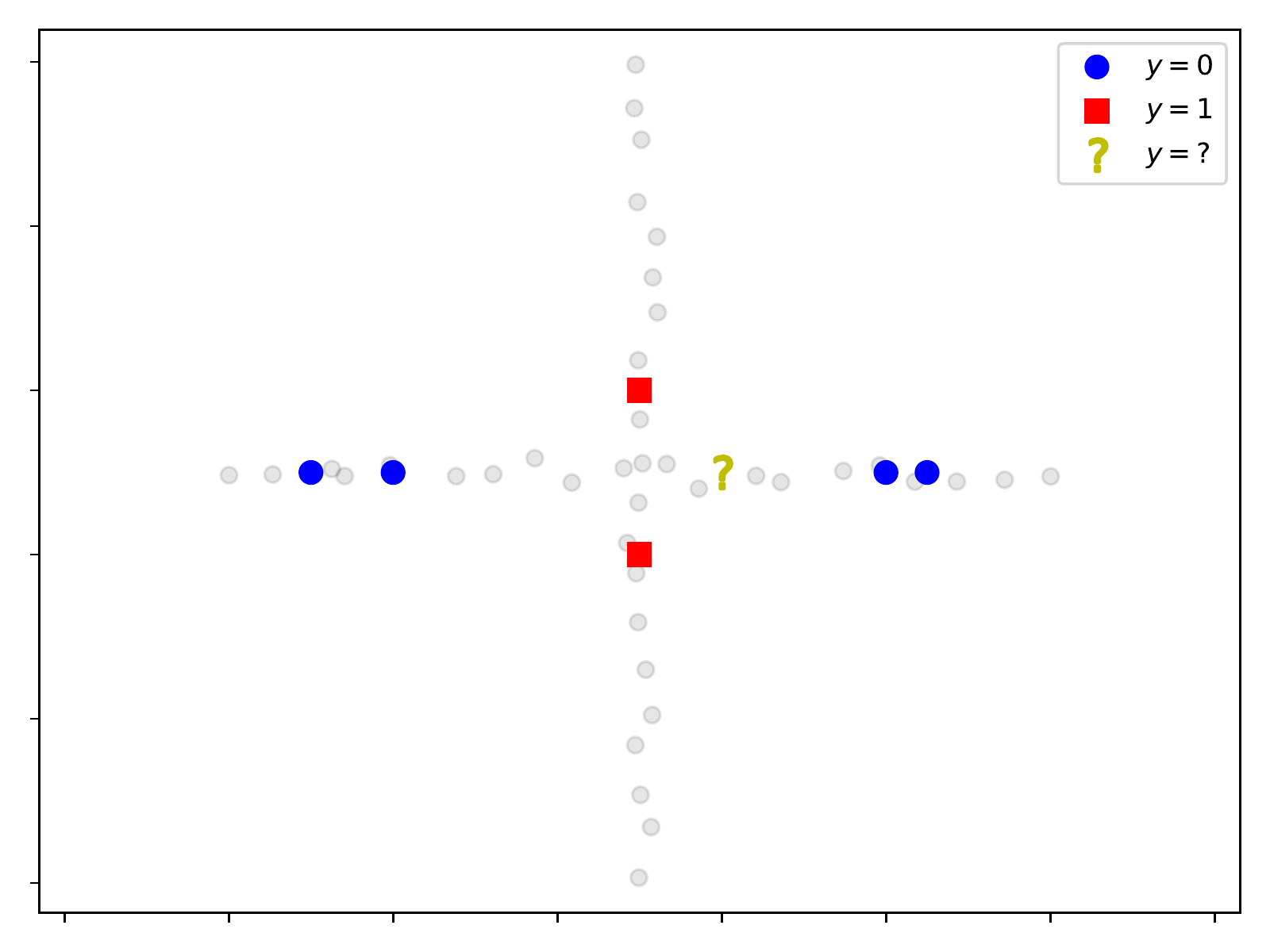}
	
	\caption{Two illustrations of the utility of unlabelled data points in semi-supervised learning: A model is asked to produce a decision for a particular test point (shown in {\color{yellow}yellow}), having observed many points, only a small number of which are labelled (shown in {class $1$ and $0$ shown in \color{red}red} and {\color{blue}blue}, respectively). A semi-supervised method makes use of the unlabelled points to deduce a dense area per class (as would be appropriate in the example on the left), or a manifold (as appropriate on the right; a linear manifold in this particular example).} 
	\label{fig:toy_data}
\end{figure}

The caveat is that semi-supervised learning methods make strong assumptions about the data or the model~\cite{SSLSurvey,van2020survey}. 
% link: assumptions of SSL, direct quote from Engelen and Hoos on 2.1 assumptions of SSL
For example, one can assume a common underlying density across points belonging to a single class, or a common manifold underlying the points of each class. Figure~\ref{fig:toy_data} illustrates two such examples. Deciding which class to assign the test data point is relatively intuitive looking at all data points, but is not clear when considering labelled data points. This highlights precisely the advantages of using the unlabelled points.

% \todo[inline]{Maciej: I think Fig. \ref{fig:toy_data} and its discussion could perhaps go to Motivation - a new subsection of 2.1, preceding challenges. This is where we could describe the intuition behind SSL. By the way: we could imagine one more toy example showing the motivation for active learning in such a subsection (assuming we want to include active learning in the survey, which is something I am not sure of).}
% \todo[inline]{Heitor: The `similar' problems were moved to that initial section. Active learning is not in the scope}

% Missing: How other authors organize existing works on SSL? The relationship between SSL, active learning and transductive learning should not be discussed in here! It is already discussed previously.  

Zhou and Li \cite{zhou2010semi} organize techniques that leverage unlabelled data in roughly three categories: semi-supervised learning, active learning and transductive learning. This high-level organization does not take into account the constraints and objectives of the learning problem, for example, active learning is not applicable if an oracle is not feasible. 
More recently, Engelen and Hoos \cite{van2020survey} organized techniques first in two classes, transductive and inductive. The majority of the techniques fall under the inductive category, since similarly to active learning, transductive learning assumes specific characteristics of the learning problem as discussed in Section \ref{sec:related_problems}. Engelen and Hoos further divided inductive techniques into wrapper methods, unsupervised preprocessing, and intrinsically SSL. Wrapper methods includes those that leverage existing supervised learning algorithms, such as co-training, self-training and boosting algorithms. Unsupervised preprocessing denotes techniques that seek to improve the performance by extracting useful features from the input data without relying on labelled data. Finally, intrinsically SSL techniques include methods that are direct extensions of supervised learning methods, i.e., they extend the objective function of the supervised method to account for unlabelled data. 

Even though it makes sense to leverage unlabelled data for the application of machine learning to streaming sources, this practice is relatively recent compared to similar approaches applied to static data. Therefore, some methodologies explored in the previously mentioned taxonomies are under-represented. For example, only a few works explore transductive learning for data streams, noticeable \cite{ditzler2012transductive}. This section focuses on inductive methods, further categorizing such methods as: Intrinsically SSL, Self-Training, Learning by Disagreement, Representation Learning, and Unsupervised and SSL Drift Detection. 
All these methods categories can be found in the batch literature, except for drift detection. Drift detection methods are of extreme importance when dealing with streaming data as unsupervised or semi-supervised drift detection can serve several purposes, such as indicating when to acquire new labels, signal relevant changes to the domain that might have not yet influenced the decision boundary, and others. 
% In this section, we discuss how SSL has been applied to data streams focusing on inductive learning, further categorizing such methods into: .

\subsection{Intrinsically SSL} \label{sec:generative}
% \todo[inline]{Jesse: Actually I don't understand why have this subsection -- it doesn't explain how to use generative models in the context of semi-supervised learning. At least it is not clear to me. Since Naive Bayes is a generative model, commonly used in data streams, this would make an excellent example to discuss (but I don't see it myself from the following).
% Heitor: The generative models are somewhat better equipped to deal with `missing' data. }

% Following the taxonomy introduced in ~\cite{van2020survey}, 
We start our characterization by discussing streaming learners that exploit the unlabelled instances directly as part of objective function or optimization procedure. 

\textbf{Maximum-margin}. Support Vector Machines (SVMs) are a popular method for supervised machine learning, based on the hypothetical maximum-margin classifier. The maximum-margin classifier's goal is to separate a binary classification problem, such that the hyperplane that splits the input space maximizes the margin. The margin is the perpendicular distance from the line to the closest points, namely the support vectors. In the construction of the classifier, only the support vectors are relevant. 
When learning a fully supervised SVM, the hyperplane is learned from fully labelled training data through an optimization procedure that maximizes the margin. In complex data scenarios, it is unlikely that there is any hyperplane that entirely separates the data. Thus the constrain of obtaining the maximum-margin is relaxed, and the goal is to find a soft-margin classifier, i.e. one that allows some of the training examples to violate the maximum-margin assumption. 

In the streaming setting, SVMs were explored for both supervised and semi-supervised problems. 
SVMs require considerable computational resources when trained on large volumes of data as the training problem involves the solution of a quadratic programming problem. \citet{domeniconi2001incremental} compared several incremental techniques for constructing SVMs and shown that they achieve predictive performance closer to the batch version while requiring less training time. Differently, from prior incremental strategies for constructing SVMs, the techniques presented in \citep{domeniconi2001incremental} inspect data only once, which is more suitable for data stream processing than other techniques. 
\citet{zhang2009mining} introduces a relational k-means based transfer semi-supervised SVM learning framework (RK-TS3VM), where instances are organized into four types: labelled (type I) and unlabelled (type III) from the same distribution as the data arriving shortly; and labelled (type II) and unlabelled (type IV) from a similar distribution to the data arriving shortly. 
% Learning from each type
Learning from type I instances follows the traditional approach to maximize the margin given labelled instances, thus solving a constrained convex optimization problem. To learn from type II, III and IV, the authors had to modify the objective function (type II and III) and rely on a relational k-means (RK) model to build new features for the labelled examples using information extracted from type IV instances. 
%% Heitor: Removed this part because it is too specific. I would need to bring the objective functions here and this can become too intricate. 
%% The knowledge from type II instances can be transferred to type I by assuming a two-task learning procedure with parameters $C_1$ and $C_2$, which controls the penalties in each task. 
%% Notice that learning from type III instances is challenging as they are unlabelled, still, it is possible to adapt the 
Empirical results presented in \citep{zhang2009mining} shows that RK-TS3VM outperform fully supervised SVM models trained only on the labelled data as well as S3VMs (semi-supervised support vector machines)~\citep{bennett1999semi}. 

\textbf{Generative models}. Generative models hypothesizes a model $p(x, y) = p(y) \times p(x|y)$ where $p(x|y)$ is an identifiable mixture distribution. Using the Expectation-Maximization (EM) algorithm~\citep{dempster1977maximum}, the mixture components can be identified given a large amount of unlabelled data. However, generative models require approximately correct modelling of the joint probability $p(x, y)$. If the modelling is incorrect, the unlabelled data may hurt performance. 
In contrast to discriminative models that only aim to estimate the conditional probability $p(y|x)$, $p(x, y)$ is more complicated to capture and too much/too little effort may lead to an incorrect model. 
%Hence, mixture components are prone to model misspecification, a detrimental factor to the accuracy of the classification. 
% Moreover, even if the modeling is correct, % the EM algorithm is prone to local minima. 
% if a local maximum is far from the global maximum, unlabelled data may hurt learning while using the EM algorithm. 
Moreover, even if the modelling is correct, unlabelled data may hurt learning if a local maximum is far from the global maximum while using the EM algorithm. 

Besides these limitations, the EM can be very slow to compute, especially when the data's dimensionality is high. Therefore, it is unlikely to apply the original EM algorithm to streaming data. \citet{cappe2009line} introduced an online version of the EM algorithm with provably optimal convergence behaviour. The online EM algorithm~\citep{cappe2009line} is analogous to the standard batch EM algorithm, which facilitates its implementation. 
% Another issue is that if the assumption of the generative model is wrong, then you end up with several incorrectly pseudo-labelled instances. Some domain knowledge is needed when deciding which generative model to use for a specific problem. More on that available in \citep{zhou2010semi}

\citet{nigam2006semi} investigated the application of EM to text classification, such that the text documents are represented using a bag-of-words (BoW) model. 
Even though a BoW representation may conceal much of the complexities of written text, the authors show that there is a positive correlation between the generative model probability and the classification accuracy for some domains. In these cases, the application of EM alongside Naive Bayes suffices to leverage predictive performance. 
Such an approach could be adapted to data streams, given the combination of the online EM method~\citep{cappe2009line} and the natural adaptation of Naive Bayes to perform incremental updates. 

%Grandvalet et Bengio ~~ Missing a clear connection to data streams! 
To compensate for the drawbacks of generative models, \citet{EntropyRegularization} proposed a method called \textit{Entropy Regularization}, aiming to only learn from unlabelled data that are informative, that is, when classes are well apart to favour the low-density separation assumption. \citet{EntropyRegularization} argue that unlabelled data are not always beneficial, mostly when class overlap occurs, and its informativeness should be encoded as a prior to modify the estimation process. The strength of the prior is controlled by a tuning parameter $\lambda$. A deterministic annealing process helps driving the decision boundary away from unlabelled data, thus avoiding poor local minima. Ultimately, the method they propose estimates the posterior distribution by maximising the likelihood of the parameters based on labelled data and at the same time being regularised by unlabelled data via $\lambda$.

\subsection{Self-training}
\label{sec:self_training}

\textit{Self-training} figures as another commonly used technique for semi-supervised learning. The idea is to let a classifier learn from its previous mistakes and try to reinforce itself. Self-training acts as a wrapper algorithm that takes any arbitrary classifier. Therefore, if we have an existing, fully-supervised learner that is complicated and hard to modify, self-training is an approach worth considering. Self-training has seen its application in natural language processing tasks such as word sense disambiguation \citep{YAROWSKY_1995} and sentiment analysis \citep{Maeireizo2004}.

In an offline scenario, self-training works as follows. Given a dataset S that consists of a set of labelled data $X_L$ and unlabelled data $X_U$ such that $S = X_L \cup X_U$, a classifier $C$ is trained on $X_L$ and after that used to predict the labels in $X_U$. The predictions with a high confidence score are assumed true and added to $X_L$ as new labelled data. The process repeats until convergence. When implementing a self-training algorithm, we must ponder the following issues: (i) how to evaluate the confidence of a prediction, and (ii) what the threshold for a "high" confidence score is? These issues remain relevant in an online scenario.
Additionally, the learner must be adapted to learn incrementally from labelled and unlabelled instances coming from the stream.

\citet{wei2006semi} introduced experiments using a self-training (i.e. self-labelling) approach for time series classification. Special considerations were taken into account to leverage a one-nearest-neighbour classifier by using unlabelled data. The main challenge in adopting such a strategy to a streaming scenario is that it requires multiple passes over the input data. More recently, \citet{jawed2020self} proposed a semi-supervised time series classification algorithm that leverages features learned from the self-supervised task on unlabelled data. It exploits the unlabelled training data with a forecasting task which provides a strong surrogate supervision signal for feature learning. 

\citet{le2019semi} proposed a self-training learner designed to receive as input either a single instance or a batch of instances at a time. A distance-based score was proposed to overcome the fact that some classifiers are unable to produce confidence scores. The confidence threshold that determines whether instances are used for self-training could be fixed or adaptive concerning the average confidence scores observed in a window. \citet{le2019semi} observed that the variant using a windowed input, distance-based scoring, and fixed confidence threshold achieves the best performance. Similarly to \citep{le2019semi}, \citet{khezri2020stds} uses a the self-training approach which uses streaming classifiers predictions along with distance-based methods to select a set of high-confidence predictions for the unlabeled data. 

% \citet{le2019semi} proposed a self-training learner with different training options. The method was designed to receive as input either a single instance or a batch of instances at a time. Second, the confidence score is computed by the learner itself, therefore the learner should be capable of yieding such scores. For example, the fraction of samples of the same label in a leaf of a Decision Tree could be used to calculate the confidence of a prediction. Another approach to calculate the confidence is a distance-based score, such that the distance from the unlabelled instance with the predicted label $\hat{y}$ is calculated to all other instances sharing the same label $\hat{y}$. Third, the confidence threshold can be fixed or adaptive with respect to the average confidence scores observed in a window. \citet{le2019semi} observed that the variant using batch input, distance-based scoring, and fixed threshold achieves the best performance.

\subsection{Learning by disagreement}
Learning by disagreement incorporates several strategies, which takes the form of learners ``teaching'' other learners. 
The canonical example is co-training~\citep{SS_COTRAINING_SEMINAL_1998}, in which two models are trained on two different views of the same data. Multi-view learning~\citep{xu2013survey} generalizes co-training to more situations where more than two views are available. 
% Co-training requires two different views of the same data,  which might not always be true. 
Also, if multiple views are not available, one approach is to enforce the disagreement among the learners' predictions~\citep{zhou2010semi}. This artificially simulates multiple views from single-view data. The disagreement among learners can be achieved through many diversity-inducing techniques, such as bootstrapping aggregation~\citep{breiman1996bagging}. 

% \textbf{Co-training}. introduced co-training, relying on the intuition of using two separate learners to ``guide" each other with the predicted labels they are most confident of during the learning process. Co-training makes a strong assumption about feature independence: the features are assumed to be divisible into two subsets such that each sub-feature set is sufficiently good to train a separate model; the two sets must also be independent given the class. Each classifier is then trained on its subset of features, gives the predictions for the unlabelled instances, and uses the predictions it is most confident of to teach the other. At the end of the iteration, one classifier receives additional labelled training examples from the other, and the process repeats. Co-training seeks for the mutual agreement between two learners. \textit{Multiview leaning} is a generalization of co-training, where more than two views are taken into account. 
\textbf{Co-training}. \citet{SS_COTRAINING_SEMINAL_1998} introduced co-training, relying on the intuition of using two separate learners to ``guide" each other with the predicted labels they are most confident of during the learning process. 
To achieve good predictive performance, co-training relies on two assumptions~\citep{SS_COTRAINING_SEMINAL_1998, ling2009does}: the \textit{consistency assumption} and the \textit{independence assumption}. 
\textit{Consistency} implies that the instance distribution is compatible with the target function, i.e. for most instances, the target functions over each view yield the same class label. 
% implies that each view individually (and combined) are sufficient to predict the class label flawlessly. 
Furthermore, the two views must be \textit{conditionally independent}, given the class label. Given two views  $X^A$ and $X^B$ and class label $y$, $X^A$ and $X^B$ are conditionally independent given class $y$ iff, given any value of $y$, the probability distribution of $X^A$ is the same for all values of $X^B$, and the probability distribution of $X^B$ is the same for all values of $X^A$.
The batch co-training procedure is relatively simple. First, learners $A$ and $B$ are trained on labelled views $X^A$ and $X^B$, respectively. Second, alternately $A$ and $B$ yield predictions for the unlabelled data. Third, the most confident predictions produced by $A$ are added to the training set of $B$ and vice-versa. This process repeats until reaching a stopping criterion. 
The first apparent challenge in applying co-training to streaming data is that it is impractical to repeat the process iteratively. Another issue that arises in an evolving streaming setting is that each view's underlying data may change over time. During periods of change, learners will likely yield incorrect predictions with high confidence contributing to their counterparts' predictive performance degradation. 
% \todo[inline]{Heitor: missing multi-view learning! Mention the survey in multi-view, then discuss what to do from there.}

\textbf{Learning by disagreement}. The key ideas behind learning by disagreement is to generate multiple learners; let them collaborate to exploit the unlabelled data; and maintain a large disagreement between them. Learning by disagreement comprehends methods such as Tri-training~\citep{zhou2005tri} and Co-Forest~\citep{li2007improve}, and it can be considered a generalization of co-training~\citep{SS_COTRAINING_SEMINAL_1998}. In these strategies, an set of learners (or ensemble) is trained on single view data, and the different views are simulated by enforcing diversity with respect to predictions through known techniques, such as bootstrapping~\citep{breiman1996bagging} or random subsets of features, as in Random Forest~\citep{RANDOM_FORESTS}. One attractive idea behind using ensembles is that the majority of the methods do not restrict which base learner should be used, thus it becomes a fairly general technique for leveraging unlabelled data. The downside, as with similar techniques, is that if not properly regularized the base learners may converge to the same decisions, as members of the ensemble train each other. 

% Ensemble for data streams
Ensemble methods are popular approaches for supervised tasks involving data streams~\cite{ENSEMBLE_DS_SURVEY_2017}, for several reasons: (1) ensembles can leverage predictive performance, often surpassing what is achievable with a single (complex) learner; (2) ensemble-based methods can be coupled with concept drift detection~\cite{LEVERAGING_BAGGING_2010,ARF,SRP}. 
% The main issue with ensembles for data streams
However, ensemble methods are costly in terms of computational resources, which can be a major concern when dealing with data streams. Even though, several ensemble methods are embarrassingly parallel, streaming and parallel implementations of such algorithms requires extra efforts to better exploit the distributed setting~\cite{diego-bd,ARF,HPCC}.

% data streams
The extension of co-training and learning by disagreement for data streams is not trivial as the algorithms that implement such techniques relies on multiple passes over the training data. 
% Unsure whether this is learning by disagreement or self-training
\citet{soemers2018adapting} leverages SSL in an unusual way while training an incremental regression tree, i.e. FIMT-DD~\cite{ikonomovska2011learning}). In \citep{soemers2018adapting} the goal is to use FIMT-DD to cluster credit card transactions, and then apply a contextual multi-armed bandit (CMAB) algorithm that makes use of the structure of the FIMT-DD model. SSL is used to assist in the training of the FIMT-DD model. Since the FIMT-DD does not split a leaf node until a sufficiently large number of instances have been observed, the number of instances at each leaf can be large, which adversely affects the CMAB algorithm. To circumvent such problem, in regular intervals, logistic regression models are used to predict the labels of the transactions at the leaves, such predictions are then presented to the tree as true labels, which then can lead to further tree splits. This approach can be viewed, even though not explicitly mentioned by the authors, as a particular case of learning by disagreement. 

\subsection{Representation Learning}
\begin{figure}
	\centering
	\includegraphics[width=0.3\textwidth]{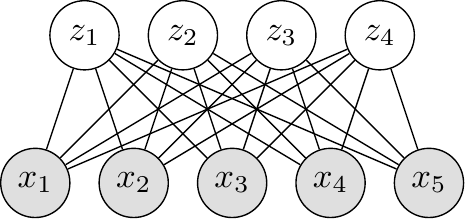} \quad\quad
	\includegraphics[width=0.3\textwidth]{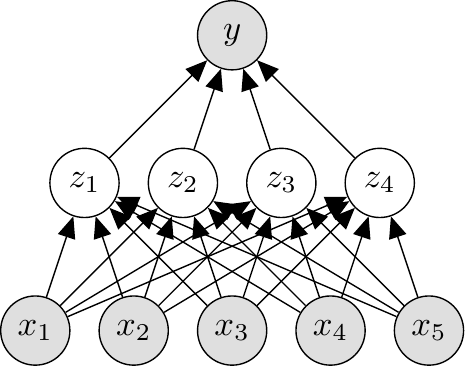}
	\caption{\label{fig:rbm}An unsupervised model (left), able to form a representation of data points as $z_1,\ldots,z_k$. In this figure an undirected graphical representation is depicted, but representations of generative models (where arrows point from $z$ to $x$) are also possible, depending on the learning algorithm chosen for this step. In a second step, the representation can then be used directly \emph{as input} to the supervised learning model (along with training label $y$, whenever it is available; i.e., learns to map $\mathbf{z} \mapsto y$). A second specific option is to consider the representation \emph{part of} a neural network (shown here, right -- where arrows show the direction of the forward pass) and use a backward pass through all layers whenever a training label $y$ is available -- thereby fine-tuning the representation for discriminative power. Shaded nodes are those observed in the data stream and white nodes are the latent/hidden representation that is learned.} 
\end{figure}

A general strategy for semi-supervised learning is to use unlabelled examples to build a representation of the input data, and then use this representation as input to a model for obtaining predictions. This technique is sometimes referred to as feature learning~\citep{RepresentationLearning}. The idea is that an improved representation will lead to improved predictions; and since representation learning can naturally be an unsupervised task, training labels are not required. Figure~\ref{fig:rbm} shows an illustration of this strategy. 

\textbf{Restricted Boltzmann machines (RBMs)} are an example of kind of model that has been used in semi-supervised data stream contexts \citep{SAC2015}. Trained using contrastive divergence, a single iteration can be carried out per instance, thus making them suitable for streams. As in the general strategy of representation learning, it is assumed that this representation improves the learning and prediction process whenever training labels are available, or predictions required, respectively. 

% Note: need to check notation with Problem Definition later
One can use the incrementally-learned representation $\mathbf{z}$ as input to any off-the-shelf data-streams classifier (naive Bayes, Hoeffding tree, etc.). 
A second option is to use the RBM's weights as the first layer of a neural network, to then be fine-tuned with back propagation \citep{DBNbp} whenever a training label is available, with some form of stochastic gradient descent; a natural incremental algorithm. Predictions are carried out via a forward pass as in any multi-layer neural network.

In RBMs the variables are binary, $z_j \in \{0,1\}$ but one may also use the probabilistic interpretations $[P(z_1|\mathbf{x}),\ldots,P(z_k|\mathbf{x})]$ as the representation for an instance $\mathbf{x}$. 

In a multi-label context, one may also obtain a representation of the label vector $y$ in a related manner \cite{ADIOS} although to our knowledge streaming variations have not yet been developed. 

\textbf{Auto-encoders} are another suitable (and related) approach. An auto-encoder is a neural network that learns to predict its own input. However, usually only the inner layer representation ($\mathbf{z}$) is of interest (hence, one can view Figure~\ref{fig:rbm} (left) as an auto-encoder with the top part of the network removed). Again, as a neural network, gradient-descent based method, learning can be an inherently incremental process. This, as well as their non-linearities, make them more suitable and powerful for streams than linear methods such as principal components analysis \citep{RepresentationLearning}. 

It can easily be argued that RBMs are a particular kind of auto encoder. In both cases, it can be emphasised that many-layer (i.e., deep) models can be used (deep representation learning). In the case of RBMs, this is typically (but not always) done greedily. In a standard auto-encoder, it is simply a deep neural network where a single layer $\mathbf{z}$ is taken. Again: the layer can be taken and given to any off-the-shelf data-stream learner (i.e., as a meta method), or turned into an instance- or batch-incremental neural network allowing back-propagation whenever labelled examples are provided by the stream. 

\begin{figure}[h]
	\centering
	\includegraphics[trim=250 110 250 110,clip=true, scale=0.45]{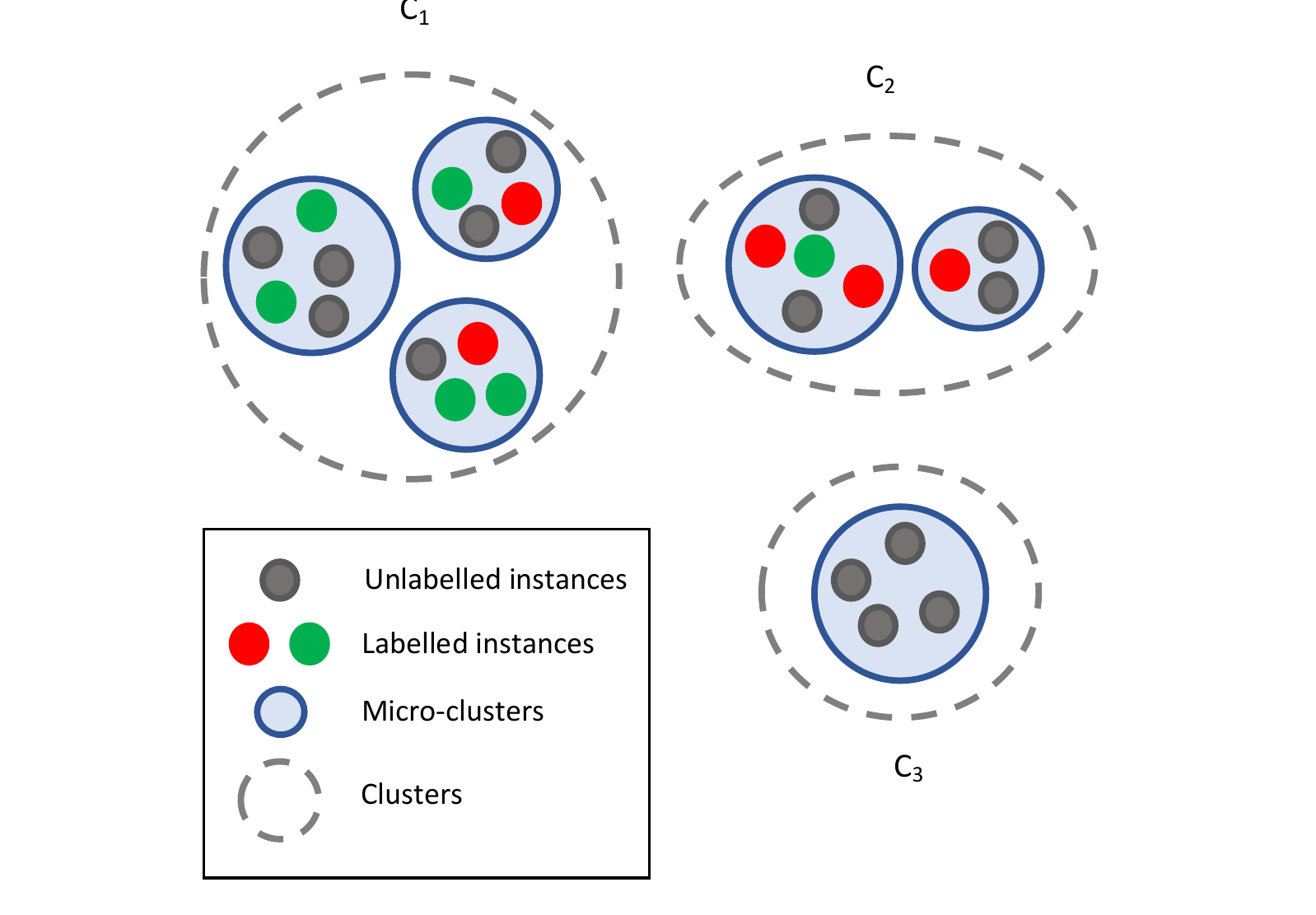}
	\caption{\label{fig:ssl_clusters} An example of clusters, micro-clusters and instances. $C_1$ and $C_2$ clusters contain a majority of instances belonging to the {\color{green} green} and {\color{red} red} class labels, respectively. $C_3$ has no labelled instances; thus no inference about the label of new instances assigned to it can be made.} 
\end{figure}

% Last minute changes to cluster representations
\textbf{Cluster representations} are useful to identify cohesive sets of input instances, which in turn can be exploited by an SSL algorithm. The cluster-then-label technique assumes that instances belonging to the same cluster may share the same label. Applying classic clustering algorithms, such as k-means, to streaming data is challenging as such algorithms repeatedly iterate over the data. The majority of the stream clustering methods incrementally update micro-clusters (summarised representations of the input data). The actual clustering algorithm is only occasionally executed in an offline step using the micro-clusters as input. Fig. \ref{fig:ssl_clusters} illustrates a situation with three clusters summarising several micro-clusters and their respective instances. The instances in Fig. ~\ref{fig:ssl_clusters} are just for illustration purposes; the whole meaning of using micro-clusters is not to store the actual instances after the micro-cluster is updated.

One such clustering algorithm to follow this approach is CluStream~\citep{CLUSTREAM}. CluStream takes a fixed number of micro-clusters $m$, which are updated whenever a new instance arrives. The offline phase of CluStream employs k-means to the micro-clusters. 
Recently, \citet{le2019semi} proposed a cluster-then-label approach utilizing CluStream, such that each cluster had an associated class label frequency counter. Pseudo-labels were assigned to arriving unlabelled data according to the most frequent label associated with its closest cluster. 
A similar strategy was earlier explored by \citet{SS_CLUSTER_DS_MASUD_2008}, where an individual model created $K$ micro-clusters from a chunk of data. The prediction was given after determining the closest $k$ nearest clusters from it. The predicted class label was the one with the highest frequency of labelled data across all of the closest $k$ clusters.

% {\color{red}
% \textbf{Cluster representations} are useful to identify cohesive sets of input instances. The basic assumption is that instances that belong to the same cluster must share the same label. There exist incremental clustering algorithms for data streams, such as ~\citep{CSLDS}.

% \textbf{Cluster representations} are also suitable in terms of implementation in many cases, as there exist incremental clustering algorithms for data streams, as in for example, \citep{CSLDS}. In hard-clustering (such as k-Means) only one of the labels $z_1,\ldots,z_k$ is set to $1$ (indicating the relevant cluster). For example, $\mathbf{z} = [1,0,0]$ may be the representation for a given instance; corresponding to an assignment to the first of three clusters. \textbf{Mixture Models} such as Gaussian Mixture Models (GMMs) are probabilistic clustering models which can give a richer representation than k-Means. The latent nodes $z_j$ of a GMM give the probability that data point $\mathbf{x}$ was generated from from cluster $j$; hence $z_1,\ldots,z_k$ is a distribution. 
% }

% ONLINE_CLUSTERING_ANOMALY_2018

% Heitor: I left it as the conclusion to the subsection on Representation learning

Realistically, any unsupervised method that can produce a useful representation of the (unlabelled) data can be considered potentially useful in the semi-supervised settings. And any algorithm for such a representation that may be suitable for a data stream is thus suitable for semi-supervised learning in a data-stream setting. Mixture models are typically trained using the EM algorithm,
which is an iterative algorithm requiring several sweeps over the data,
% Perhaps I will remove the explanation about EM from here, since it can be explained earlier in the intrinsically SSL subsection. For now, let's keep it in here. 
however it can be adapted to streams \citep{cappe2009line}. In fact, the EM algorithm and k-means are special cases of self-training (see Section~\ref{sec:self_training}).
% Heitor: I added a reference to an online EM algorithm by Cappe

\subsection{Unsupervised and SSL Drift Detection}
% Introduction
Real-world problems tend to be very dynamic. For example, consumer behaviour may change as time goes by, a group of people can change their opinion about a product or a political party, the attacks a network receives change as new barriers are created, and so on. Learning from data that distribution may change over time is challenging for conventional batch machine learning algorithms. These algorithms assume that the data distribution is static. 
Conventionally, data streams that contain drifts are identified as evolving streams. 

% Different aspects to consider...
There are many aspects to consider when discussing concept drift, including its cause, rate, and data distribution. Generally, a drift can be characterized either as ``real'' or ``virtual''~\citep{SURVEY_CD_GAMA_2014}. A real concept drift happens when changes affect the class labels' posterior probabilities, $p(y|X)$, i.e., the output variable distribution changes affecting the upcoming predictions. Virtual concept drift is said to occur if the distribution of the incoming data $p(X)$ changes without affecting $p(y|X)$. Usually, there is not much interest in virtual drifts because they do not change the output's conditional distribution. 
A sizeable amount of research has been dedicated to discuss different aspects concerning concept drift ~\citep{CHAR_CD_WEBB_2016,SURVEY_CD_GAMA_2014}. This section focuses on discussing concept drift in scenarios where labels are delayed and often partially available. 

% \todo[inline]{Heitor: Mention APT in here!}

% Delayed, supervised and unsupervised drift detection
Most concept drift detection algorithms are applied to the univariate stream of correct/incorrect classifier predictions. 
Such strategies require that labelled data is available as soon as possible to respond to concept drifts in a timely fashion. From a practical standpoint, despite their intrinsic differences, most drift detectors trigger when the observed model's predictive performance starts to degrade. 
Algorithms such as ADWIN~\citep{ADWIN:2007} and EDDM~\citep{EDDM:2006}, were tested under the assumption that labelled data is available almost immediately. However, if the ground-truth is not immediately available, then these algorithms' ability to timely detect drifts is severely decreased.

\begin{figure}
    
	\includegraphics[trim=10 10 10 10, scale=0.45]{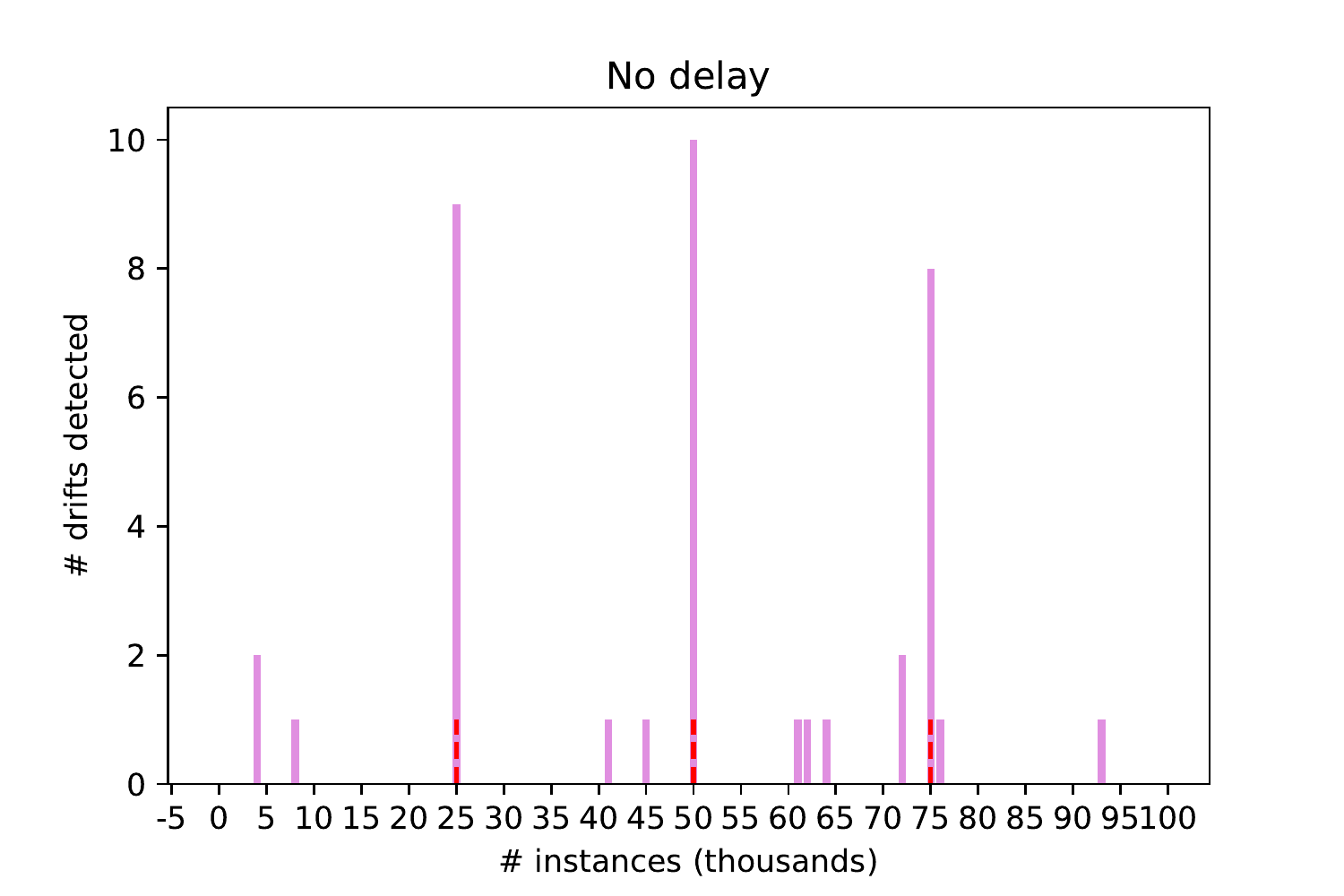}
	\includegraphics[trim=10 10 10 10, scale=0.45]{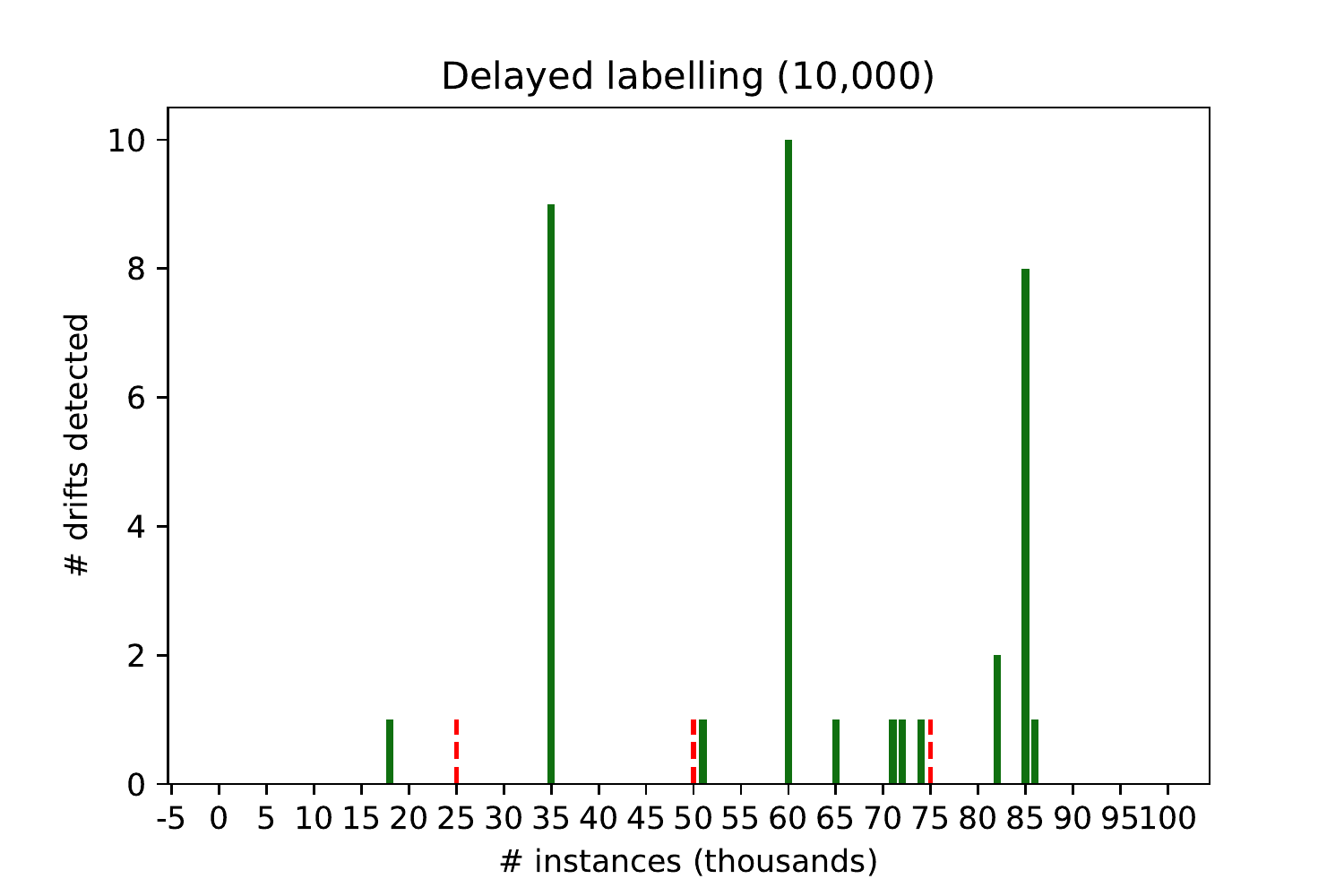}
	
	\caption{Drifts detected by a 10 learner SRP model using ADWIN on AGRAWAL with and without labelling delay. Red dotted vertical lines indicate the location of concept drifts.} 
	\label{fig:SRP_ADWIN_drifts}
\end{figure}

\begin{figure}
    
	\includegraphics[trim=10 10 10 10, scale=0.45]{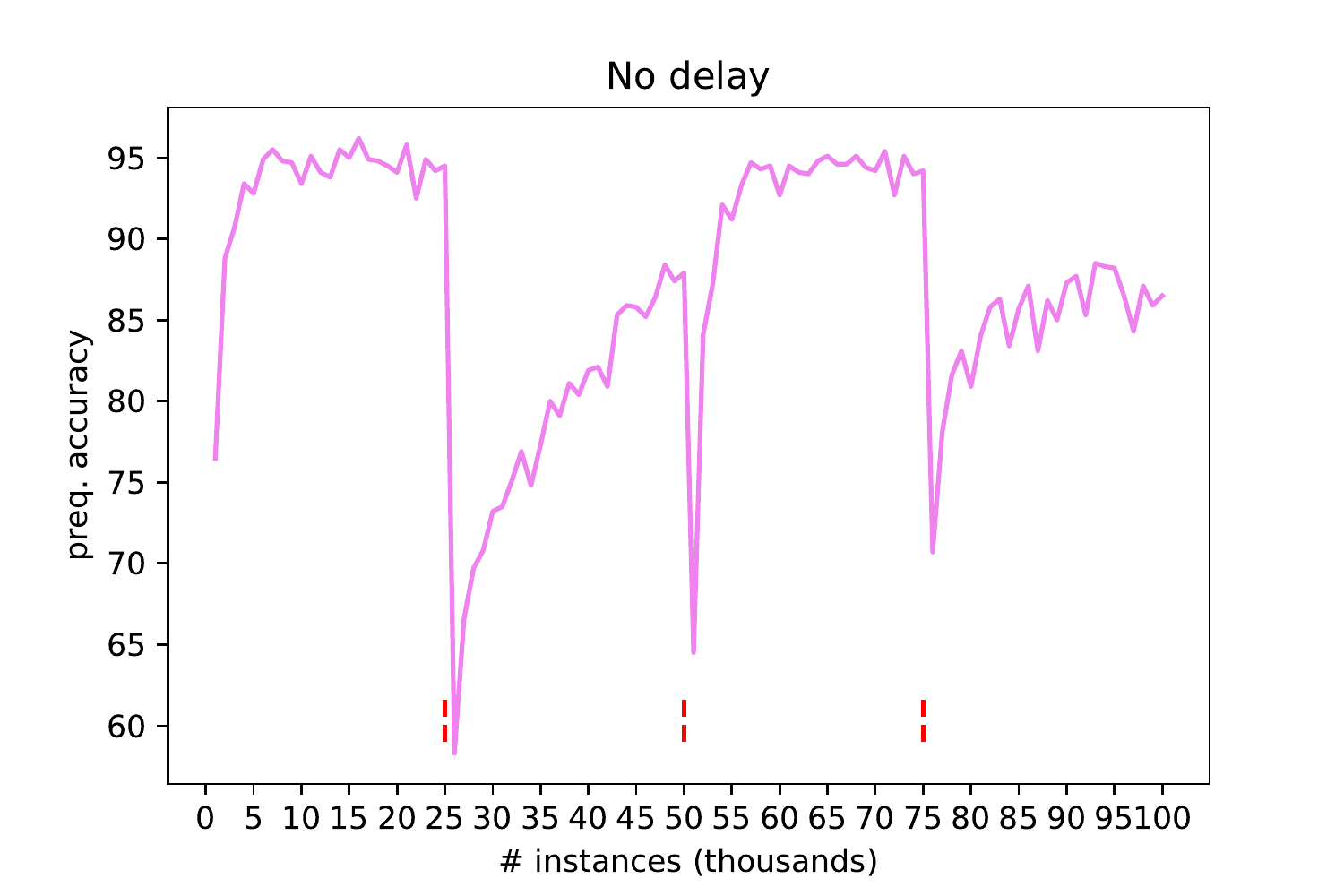}
	\includegraphics[trim=10 10 10 10, scale=0.45]{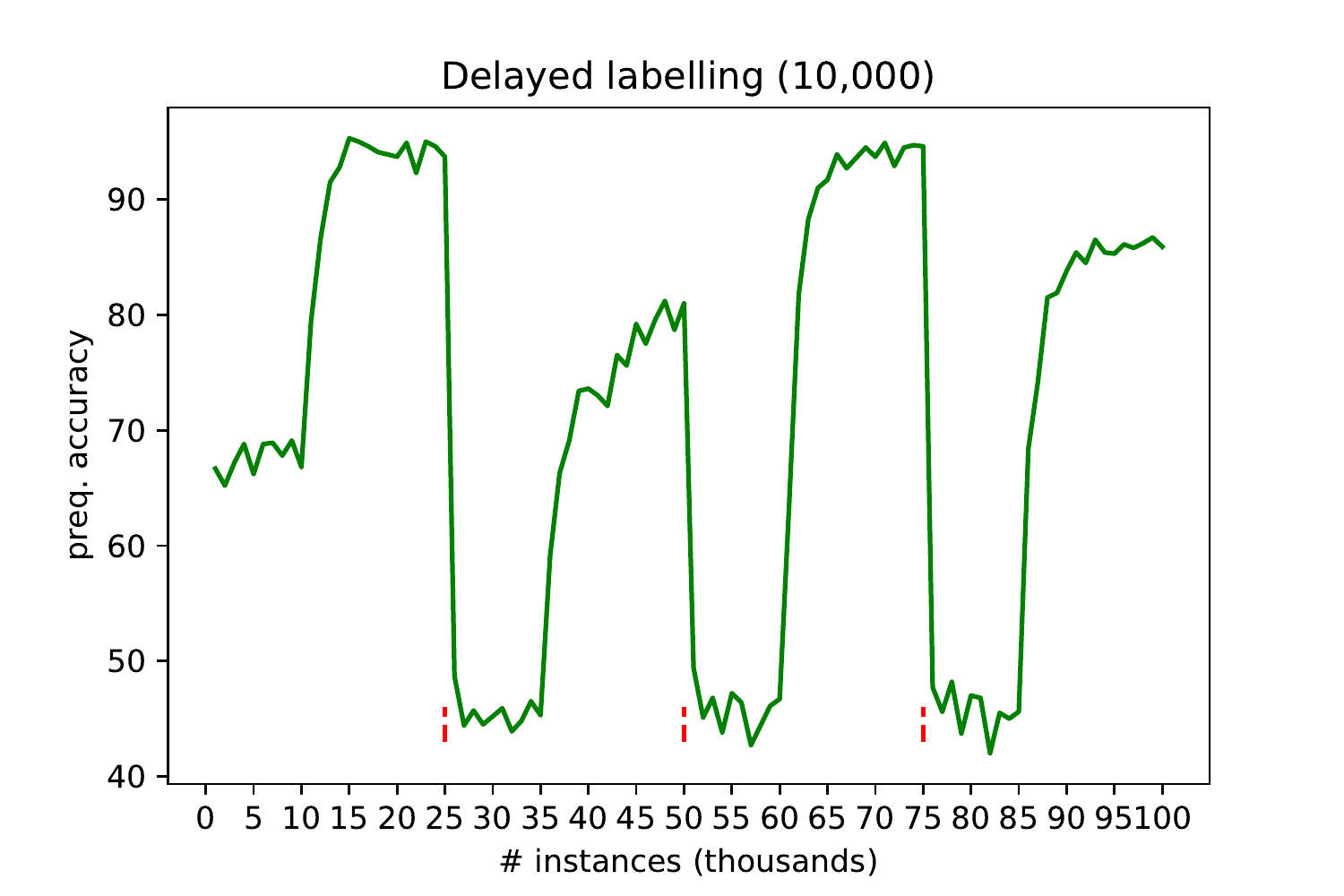}
	
	\caption{Accuracy by a 10 learner SRP model using ADWIN on AGRAWAL with and without labelling delay. Red dotted vertical lines indicate the location of concept drifts.} 
	\label{fig:SRP_ADWIN_acc}
\end{figure}

% Experiments to illustrate the impact of delay
To illustrate the impact of delayed labelling on a drifting stream, we present a small experiment with data generated using the AGRAWAL generator with 3 abrupt concept drifts (at instances $25,000$, $50,000$, and $75,000$). In these experiments, we used an ensemble algorithm~\cite{SRP} capable of detecting and adapting to changes by resetting base models whenever changes are detected (by ADWIN~\citep{ADWIN:2007}) on their univariate stream of correct/incorrect predictions. Figure \ref{fig:SRP_ADWIN_drifts} depicts the amount of concept drifts detected (y-axis) over the processing of $100,000$ instances with and without delayed labelling. In Figure \ref{fig:SRP_ADWIN_drifts}, the detections for the ``No delay'' experiment shows a high rate of detection immediately after the concept drifts, except for a few arbitrary drift signals in-between the concept drifts. A labelling delay of $10,000$ instances severely impacts the ability to detect the changes, as shown in the Delayed labelling variant, where the detections still occur but with a shift of approximately $10,000$ instances. 
% Accuracy of the experiments
The impact on the predictive performance in the delayed and not delayed scenarios is clearly observable in Figure \ref{fig:SRP_ADWIN_acc}. The experiment depicts the prequential accuracy~\footnote{A window of $1,000$ instances was used in this prequential accuracy evaluation.} as the $100,000$ instances are processed. The delayed labelling causes longer periods of poor accuracy (below $50\%$ on a balanced binary problem).

The occurrence of a concept drift indicates that something is off concerning the learning model. Several actions can be taken after drift is detected, such as raise the alarm to the user, trigger automatic changes to the underlying model, or selectively request new labelled data (active learning). Research is often devoted to automatic detection and recovery~\cite{ARF}; however, these can be frowned upon because they give less control over the model. Raising the alarm and signalling the need for new labelled data is a less drastic approach that gives the data scientists behind the model more control. This control is especially useful in scenarios where prediction stability and fairness are equally (or more) important than accuracy. 
% stability
A brand new model might be accurate for the immediately new data, yet it may be unstable to produce predictions given that it was only trained on a small amount of data. 
% fairness
On top of that, automatically re-training the model leaves room for unfair treatment of underrepresented groups in the data, e.g., it is harder to prevent gender and race discrimination if the model is being updated on the fly without being reviewed. 
% Finally...
Therefore, one interesting application is to detect concept drift and only notify the user. Depending on the machine learning pipeline, a new model might be created on the fly to replace the old model, but it will not be deployed immediately without user interference. Another option is to notify the user to suggest instances for labelling using active learning strategies for data streams~\citep{vzliobaite2013active}. 

% Separate discussion: a model constantly being updated by new inputs is susceptible to adversarial attacks, since adversaries can inject instances to bias the model. 

% \citet{CD_DELAYED_ZLIOBAITE_2010} {\v{Z}}liobaite~
\citet{CD_DELAYED_ZLIOBAITE_2010} presented an analytical view of the conditions that must be met to allow concept drift detection in a delayed labelled setting. These conditions depends on characteristics of the concept drift, i.e. how the change affected the input probability $P(X)$ and the posterior probabilities $P(w_i | X)$ of the class labels $w_i$. \citet{CD_DELAYED_ZLIOBAITE_2010} thoroughly discussed when it can be expected that changes are observable or not as shown in Table \ref{tab:when_cd_detection}. One interesting connection between semi-supervised learning and unsupervised concept drift detection is that if the underlying marginal data distribution $P(X)$ over the input does not contain information about $P(y|X)$ or indicates changes on $P(y|X)$, then it is impossible to exploit unlabelled data to improve a supervised learner (SSL) or detect a change ((3) in Table~\ref{tab:when_cd_detection}). 

\begin{table}[h]

\begin{tabular}{clll}
                        &                                & \multicolumn{2}{c}{$P(w_i | X)$}                                                       \\
                        &                                & \textbf{change}                                            & \textbf{no change}                          \\ \cline{3-4} 
	\multirow{2}{*}{$P(X)$} & \multicolumn{1}{l|}{\textbf{change}}    & \multicolumn{1}{l|}{(1) important \& observable}   & \multicolumn{1}{l|}{(2) observable} \\ \cline{3-4} 
                        & \multicolumn{1}{l|}{\textbf{no change}} & \multicolumn{1}{l|}{(3) important \& unobservable} & \multicolumn{1}{l|}{(4) no drift}   \\ \cline{3-4} 
\end{tabular}
\caption{When concept drift detection is observable and important (affects the decision rules), adapted from ~\citep{CD_DELAYED_ZLIOBAITE_2010} \label{tab:when_cd_detection}}
\end{table}

% Existing algorithms for partially and delayed labelled streams. 
A handful of algorithms focus on drift detection on delayed, partially labelled or unlabelled data streams. Examples include SUN~\cite{CD_UNLABELED_WU_2012} and the method from 
% \citet{CD_SS_2001}
Klinkenberg~\citep{CD_SS_2001} based on support vector machines. The former uses a clustering algorithm to produce `concept clusters' at the leaves of an incremental decision tree, and drifts are identified according to the deviation between history concept clusters and the current clusters. 
\citet{baron2015} proposed an approach that dynamically determines the boundaries of windows by detecting significant changes in classifier confidence scores using a limited number of labelled instances. This approach is integrated with a classifier to form a complete SSL framework that utilises dynamic chunk boundaries to address concept drift and concept evolution. 
\citet{cerqueira2020unsup} presents an unsupervised drift detector based on a teacher-student approach, where a predictive model (teacher) is built using an initial batch of labelled training data. The teacher's predictions are used as class labels to train a surrogate model (student), which will learn to mimic the teacher. A drift detection algorithm is used to identify variations in the mimicking error of the student. 
The hypothesis is that if the mimicking error increases, then it means that a concept drift has occurred.

\section{Fair comparative analysis} \label{sec:comparative_analysis}

Like other machine learning methods, SSL methods should be evaluated in a realistic process to verify their capabilities while considering other applicable methods. 
\citet{van2020survey} observed that additional factors have to be considered during evaluation compared to fully supervised learning scenarios. 
% \citet{van2020survey} in their recent extensive survey on SSL observed that, compared to fully supervised learning scenarios, additional factors have to be considered during evaluation. 

First and foremost, the question arises of whether the use of a semi-supervised approach yields performance gains compared to supervised methods~\citep{REALISTIC_EVAL_SEMI_SUPERVISED_2018}. Furthermore, a comparison of an SSL method of interest with other SSL methods is required. Similarly to other machine learning methods, the selection of the data for which predictions are evaluated has to be followed by calculating performance measures. Interestingly, due to the latency of ground truth labels, multiple predictions made for a single instance at different times before the arrival of its true label may be considered in the evaluation~\citep{grzenda2019delayed}. The objective of this section is to address the unique aspects of the evaluation of semi-supervised stream mining methods while taking into account non-negligible delays in the availability of ground truth labels. 

% Changed from Preliminaries to this new title. The idea is to avoid suggestions from reviewers to move a preliminaries section to the initial sections of the paper :)
\subsection{Evaluation of machine learning methods}
%\subsubsection{Key aspects of related works}
%\todo[inline]{this is where key related works can be mentioned together with their limitations}
Evaluation of machine learning methods necessitates applying an appropriate combination of error estimation methods, which include but are not limited to the selection of data used for model development and testing. In addition, performance measures matching domain needs have to be selected. An extensive study on evaluating learning algorithms by \citet{japkowicz2011} has already provided in-depth coverage of performance measures for classification, how they should be calculated and how different candidate methods should be compared. However, the latter work is focused on the fully supervised batch learning paradigm, i.e. the use of fully labelled data sets rather than partially labelled data streams for both learning and evaluation purposes.  

In recent years, there has been an increasing number of studies on evaluating learning algorithms under other settings than fully supervised batch learning. In particular, \citet{REALISTIC_EVAL_SEMI_SUPERVISED_2018} examined the impact of various factors such as hyperparameter tuning, class imbalance, and volume of unlabelled data on the evaluation of semi-supervised deep learning methods and revealed how important careful consideration and documenting of these and other factors could be. However, in this major study on evaluating SSL methods, the focus was limited to batch learning. As far as the evaluation of machine learning methods applied to data streams is concerned, new procedures aimed at how data streams should be used for online learning and evaluation purposes were developed. A summary of the most popular of these methods was made in \citep{BOOK_MLDS_2017}. Importantly, these methods do not consider label latency.
Besides new performance measures such as measures reflecting temporal dependencies in the data \citep{pitfalls2013}, intermediate performance measures capturing the performance of multiple predictions made over time for a single instance \citep{grzenda2020ijcnn} were proposed. A particular limitation of these works from the perspective of our study is that these measures assume a fully supervised setting. 
% Removed this part as the citations are already mentioned above, so it was safe to remove and save some space :)
% In line with this assumption, a data stream is expected in \citep{pitfalls2013, grzenda2020ijcnn} to be fully labelled.

% \todo[inline]{Heitor: We could refer to the definition of delayed and partially labelled presented on section 2, like this: ``There are relatively few studies on SSL methods applied under delayed labelling settings (as defined in Section \ref{sec:problem_setting}), [...]''}
While each of these studies refers to at least one of the aspects of the evaluation of SSL methods for delayed partially labelled data streams, to the best of our knowledge, only selected aspects of the evaluation of SSL methods were covered in them. There are relatively few studies on SSL under delayed labelling settings (as defined in Section \ref{sec:problem_setting}), which we refer to below.

Taking into account the complexity of the evaluation of machine learning methods~\citep{japkowicz2011}, in this section we aim to focus on the unique aspects of the evaluation of SSL methods for partly labelled delayed data streams, rather than provide a holistic view of all aspects of the evaluation of stream mining methods. Our intention is not to repeat these aspects which are common to the evaluation of other stream mining techniques, but to pay particular attention to how the evaluation relevant for this study differs from the evaluation suggested for related tasks. Furthermore, we conclude this section with a unified view of the recommendations we make. 
% To make this possible and avoid ambiguity we propose new concepts and notation with the aim of making these concepts, some of which were used in the past with different meanings, precise. 

% \subsubsection{Data streams used for the evaluation}
% \todo[inline]{Heitor: Maybe these definitions, $\Psi$ and $T_\mathrm{min}, T_\mathrm{max}$ could be moved to section \ref{sec:problem_setting}. It is probably fine to keep them in here, but I was trying to define $T_\matchrm{\cdot}$ in terms of $T(\cdot)$ and I couldnt come up with anything yet.}
All data streams considered in this section are assumed to be delayed data streams. Furthermore, let us observe that while data streams are infinite by definition, the evaluation of learning methods inevitably relies on stream sections
$\Psi[T_\mathrm{min},T_\mathrm{max}]$.
% a sequence of instances and true labels $\Psi[T_\mathrm{min},T_\mathrm{max}]$ that arrived during a time window $[T_\mathrm{min},T_\mathrm{max}]$ or the subsets of these instances and labels. 
In particular, the period $[T_\mathrm{min},T_\mathrm{max}]$ may be the period of a single concept or may span multiple concepts. 

The remainder of this section is organised as follows. First, an overview of the evaluation processes and measures applicable to stream mining methods is presented. This is followed by a discussion of the way standard evaluation practices can be tailored to enable a comparison of SSL methods with fully supervised methods. Next, the role of data streams used in the evaluation and other key factors influencing the evaluation outcomes are discussed. This provides the basis for the unified evaluation process proposed in this study, which concludes this section.

\subsection{Evaluation of stream mining models}
\subsubsection{Evaluation process}
% \todo[inline]{I think we should briefly refer here to measures and techniques such as prequential etc. here to make it clear what is beyond the scope of this study, as we adopt it from stream mining studies done in fully supervised and/or immediate labelling setting}
% \todo[inline]{Heitor: Agreed. I will start off shamelessly copying some text that we used on the delayed evaluation paper, then improving and adapting it. }

% Important aspects when evaluating data streams

% ``HOW'' past predictions influence the predictive performance
% The evaluation of machine learning methods for data streams is often focused on \textbf{``how''} past predictions influence the value of predictive performance measure for the current model.
The evaluation of machine learning methods for data streams is often focused on \textbf{``how''} past predictions influence the current model's predictive performance measure.
% Test-then-train
The most straightforward approach is to evaluate predictions in a \textbf{test-then-train} fashion; as the name implies, each instance is first used for testing and then for training. The predictive performance of the learning algorithm in test-then-train represents the average value of all the instances assessed up to that point in time \cite{BOOK_MLDS_2017,grzenda2019delayed}. 
% Limitation of test-then-train
In test-then-train evaluation, the latest predictive performance estimation is influenced by all previous predictions. This characteristic is desirable when the goal is to understand how well the model performs up to a given point in time. However, it may not give a clear view of how well the model is performing at a given period of the stream. For example, several recent incorrect predictions (perhaps caused by a concept drift) may be shadowed by thousands of old correct predictions.

% periodic holdout
One approach to avoid undesired influence from previous predictions is to perform a \textbf{periodic holdout} evaluation, where training and testing are interleaved using predefined windows, such that one window is used for training and the following used only for testing. % Prequential
This approach can be considered wasteful as labelled data that could be used for training (after testing) is discarded. Thus, an alternative approach is to use a \textbf{prequential evaluation}~\cite{BOOK_MLDS_2017}, where test-then-train is applied to a sliding window, or a fading factor is used, to forget old predictions. A more in-depth discussion of both approaches to prequential evaluation can be found in \citet{gama2013}. Importantly, when data are only partly labelled, the prequential evaluation is still applicable, as the loss function can be calculated only for those observations for which labels $\mathit{y}_k$ are known~\cite{gama2013}.
% Cross-validation
A well-accepted approach for increasing confidence in the evaluation results of batch evaluations is cross-validation. The challenge associated with cross-validation on a streaming setting is that it is infeasible to reprocess the whole stream. To cope with this constraint \citet{EFFICIENT_EVALUATION_BIG_DS} introduced an approach for cross-validation in an online setting, where the models are trained and tested in parallel on different folds of data. Furthermore, in delayed data streams, predictions are typically made at $T(x_k)$ and can be evaluated only after the corresponding true labels arrive. This means \textbf{verification latency}~\cite{ditzler2015} occurs. 
% This kind of evaluation was performed in this way was included in \cite{ARF}.
% Importantly, all the aforementioned approaches assume that for every instance its true label is immediately available and can be used by the evaluation process with no delay. In other words, the stream for which the evaluation is made is assumed to be an immediate and fully labelled stream. 
% \todo[inline]{Maciej: @Heitor, I think we can add the sentences above to link it with the definitions we have at the beginning}

% \todo[inline]{Heitor: @Maciej, please check this paragraph. I believe I addressed all the requirements from your previous comment (see comment at the end of this paragraph). \\
% @Heitor: I think it is OK. I have added 1-2 more sentences.

% I thought about adding an image with an evaluation with $B=100$ for several algorithms and the corresponding $\Psi(T)_a$ similar to \cite{grzenda2020ijcnn}, it could be even one of the figures that we have in that paper. 
% @Heitor: I will try to come up with some diagram, ideally a new one
% }
% ``WHEN'' labels become available
\citet{grzenda2019delayed} claim that besides \textbf{``how''} predictions affect the predictive performance, it is also key to consider \textbf{``when''} labels are made available as part of the evaluation. This leads to the concept of \textbf{continuous re-evaluation} introduced in \citep{grzenda2019delayed} and further explored in \citep{grzenda2020ijcnn}. The goal of continuous re-evaluation is to observe if, and how fast models can transform an initial, possibly incorrect prediction made at $T(x_k)$ into a correct prediction before the true label arrives at $T(y_k)$.  
While waiting for a label $y_k$ to determine whether a prediction was correct or not, the model is incrementally trained with labels from other instances that arrived before $T(y_k)$. These updates to the model may change the initial prediction yield for $x_k$. 
%This kind of analysis allows for further understanding of how fast a given model can learn and correct predictions.

\begin{figure}
\centering
    % trim=left bottom right top
	\includegraphics[trim=20 100 20 90, scale=0.35]{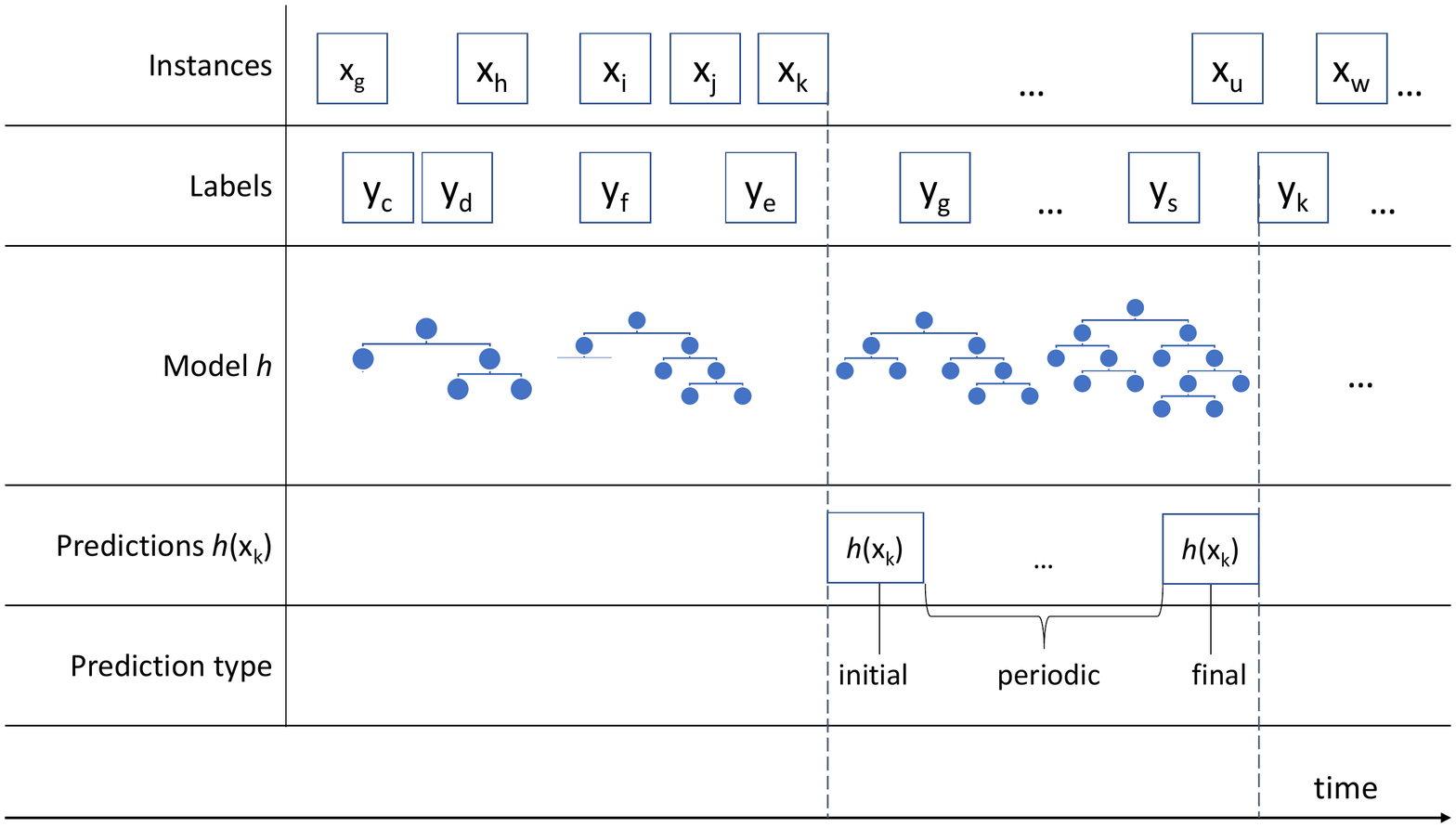}
	\caption{The types of predicted labels under delayed labelling setting. Based on \citet{grzenda2019delayed}}. 
	\label{fig:prediction_types}
\end{figure}

% Example of application
This evaluation is essential to scenarios where evolving predictions are relevant, such as continuously re-assessing whether a recently released application is  ``malware'' or ``goodware'' until the ground-truth is available~\cite{ceschin2020machine}. Continuous re-evaluation is the generalisation of the test-then-train approach, as it provides a way of calculating and assessing \textbf{initial predictions} made for $x_k$ at $T(x_k)$, possible \textbf{periodic predictions} made for the instance before $T(y_k)$, and \textbf{final predictions} made at $T(y_k)$ i.e. immediately before using the true label to possibly update a model \cite{grzenda2019delayed}.
Fig.~\ref{fig:prediction_types} illustrates the way all these types of predictions are produced for the $x_k$ instance. 
% The way all these types of predictions are produced for $x_k$ instance is shown in Fig.~\ref{fig:prediction_types}.
Continuous re-evaluation assumes that for every instance, its true label is available with non-infinite delay~\citep{grzenda2019delayed}. 
% In other words, the stream for which the evaluation is made is assumed to be a fully labelled stream. 
% Limitations and applicability
Even in scenarios where evolving predictions are not essential, it might still be useful to also consider 
% not only the initial predictions performance. The reason is that the
intermediary predictive performance as it indicates how fast the model can adapt to predictions given that other labelled data were made available. 
% This indicates how fast a model addresses instances similar to $x_k$ while waiting for $y_k$. 
% Still works with SSL. This statement is super obvious, maybe it could be improved. 
Finally, one could potentially adapt continuous re-evaluations to partially delayed labelled streamas, the main difference being that only instances where $y$ arrives will be used for assessment. 
% Such an adaptation could be used to evaluate SSL methods for partially labelled delayed data streams.
% Guide for the following sections, just to ``glue'' it together
%In the following sections, we discuss the evaluation of SSL methods

% For example, considering the predictive maintenance example, several predictions can be obtained for a given machine when trying to determine whether it is going to break or not. The sooner a break is detected, the better; if it is only discovered a few instants before the machine breaks, then it is not useful.

%   \todo[inline]{I think, this is where we could discuss initial, final and continuous re-evaluation, as these options are explicitly related to delayed labelling setting. In addition, the fact that continuous re-evaluation does not explicitly address semi-supervised settings can be mentioned}

\subsubsection{Performance measures}
A fundamental aspect of every evaluation of stream mining methods is the selection of performance measures to be calculated. Similarly to other stream mining settings, apart from measures such as accuracy, measures developed to address the unique aspects of streaming data, including possible temporal dependencies, should be considered. Kappa temporal ( $\kappa^{+}$)~\citep{pitfalls2013} is of particular importance for many data streams as it compares the performance of a model against the performance of a simple ``No Change'' model. The ``No Change'' model always predicts the next $y$ using the previous true label, which causes it to be consistently correct when temporal dependencies are expected.
A summary of all the aforementioned measures can be found in \citet{BOOK_MLDS_2017}. In addition, when multiple predictions are made for a single instance, the values of measures such as accuracy and kappa can be aggregated into intermediate  performance measures~\cite{grzenda2020ijcnn}. In this way, the performance of both initial and periodic predictions made for an instance $x_k$ before its true label arrival $T(y_k)$ can be assessed~\citep{grzenda2019delayed}. 
% The intermediate performance measures are calculated under the continuous re-evaluation process \cite{grzenda2019delayed}. 

Furthermore, it is essential to note that for the evaluation to be complete, the memory and computational overhead should be reported. In \cite{Masud2012},  running time and memory allocation were reported for both supervised and semi-supervised technique. Importantly, the evaluation of stream classification methods in \cite{ARF} revealed that some of the tested methods failed to process data streams even when 200GB of operating memory were made available for these methods. Hence, ideally, CPU and memory consumption should be reported for all evaluated methods, including  SSL methods, along with the measures focused on assessing the ability of individual methods to minimise the loss of prediction. 

Streaming algorithms are expected to be efficient, so it is reasonable to also assess them in terms of computational resources~\citep{BOOK_MLDS_2017}. On top of that, other key features of the data stream should be reported, such as the proportion of labelled and unlabelled instances, the number of true labels in the latest window, the proportion of individual classes, and others. All these factors may have a substantial impact on the performance of a stream mining method. 
% Furthermore, taking into account the results of other studies on SSL methods, not only performance measures of individual models, but also other numerical indicators revealing the key features of a data stream should be reported for individual instance windows of the data stream for which evaluation is performed. In particular, these indicators should document for individual windows the proportion of labelled vs. unlabelled instances, the number of true labels which have arrived during the instance window, and the proportion of individual classes, as all these factors may have a substantial impact on the performance of stream mining methods.

\subsection{Comparison of semi-supervised vs. supervised methods}

% \subsubsection{The evaluation scenario based on removing some true labels}
\subsubsection{Evaluation based on removing some labels} % Suggested short name for the section
When an SSL method is considered, its merits should be verified through comparison against supervised methods, including methods of possibly lower computational complexity. This should be done by using an appropriate combination of evaluation process and performance measures. Whether other periodic predictions are justified by the domain problem or not, continuous re-evaluation adapted to a partially labelled setting can be used to analyse the performance of just initial predictions, or possibly also final and periodic predictions.
As previously mentioned, computational resources are paramount to stream mining algorithms. Hence, when a supervised method yields the same performance as an SSL method, which is achieved at a lower computational overhead, it is natural that the supervised method will be preferred. SSL methods may require more computational resources as they potentially use all incoming instances for training.
%, while supervised methods may ignore unlabelled instances. This confirms the need for tracking the computational and memory needs of all the techniques under consideration. 

Comparison of an SSL method against a fully supervised method can be made in two ways. First of all, some labels can be removed from an initial data stream to provide a delayed and partially labelled data stream processed by an SSL method \cite{baron2015,le2019semi,Masud2012}. We will refer to such a data stream as a {\em reduced partially labelled data stream}, which we denote by $F_\mathrm{U}(\Psi[T_\mathrm{min},T_\mathrm{max}],p_\mathrm{u})$. We propose to generate such a data stream, by removing with probability $p_\mathrm{u}$ individual true labels  $\{(\cdot,\mathit{y}_k)\}$ from $\Psi[T_\mathrm{min},T_\mathrm{max}]$. In this way, in every run of an evaluation process fed with $\Psi[T_\mathrm{min},T_\mathrm{max}]$ data, a possibly different reduced partially labelled data stream $F_\mathrm{U}(\Psi[T_\mathrm{min},T_\mathrm{max}],p_\mathrm{u})$ will be generated and used to evaluate the impact of the reduced number of true labels on the evaluation process. Importantly, this means that each of the originally labelled instances $\{(\mathbf{x}_k,?)\}$ is converted with probability $p_\mathrm{u}$ into an unlabelled one. 

%\todo[inline]{Emphasise here the need for precise description, which is why we propose notation}
%Without the loss of generality, let us define that $F_\mathrm{U}(\Psi,p_\mathrm{u}=0,T_\mathrm{min},T_\mathrm{max})$ is the data stream $U(T_\mathrm{min},T_\mathrm{max})$.

%\todo[inline]{Define $U(T_min,T_max)$}

Furthermore, let us observe that any fully supervised method will ignore the existence of unlabelled instances. Hence, it will operate on what we call a {\em reduced fully labelled data stream} $F_\mathrm{L}(\Psi[T_\mathrm{min},T_\mathrm{max}])$. This stream refers to the one created from the initial stream after removing instances for which no labels have arrived until $T_\mathrm{max}$. In other words, $F_\mathrm{L}(\Psi[T_\mathrm{min},T_\mathrm{max}])$ neglects the existence of unlabelled instances. Hence, it provides input for fully supervised learning methods.

The practice of removing labels to create partly labelled data sets is frequently present in studies on SSL methods. \citet{van2020survey} observed that data sets used in research are usually obtained by removing several labels from existing supervised learning data sets.
In line with these practices, a comparison of the performance measures attained by a fully supervised method operated on a $F_\mathrm{L}(\Psi[T_\mathrm{min},T_\mathrm{max}])$ data stream and an SSL method operated on a  $F_\mathrm{U}\Big(F_\mathrm{L}\big(\Psi[T_\mathrm{min},T_\mathrm{max}]\big),p_\mathrm{u}\Big)$ data stream can be made. Such a comparison of SSL methods operating on partly labelled data with supervised methods using fully labelled data streams has been made {\em inter alia} in \citep{baron2015,dyer2014,le2019semi,Masud2012}.

%Interestingly, an SSL method using 95\% of labelled data has been shown to outperform fully supervised Adaptive Hoeffding tree using fully labelled data.

Moreover, the impact of the $p_\mathrm{u}$ value on the performance measures of an SSL method should be analysed to provide insight into the way the method responds to varying volumes of labelled and unlabelled data. In particular, a supervised method's performance can be compared with an SSL method's performance operating on a reduced number of labelled instances. 
It is now well established that some batch SSL algorithms may work well or not depending on the volume of labelled and unlabelled data~\citep{REALISTIC_EVAL_SEMI_SUPERVISED_2018,van2020survey}. Analysing individual methods' performance under varied ratios of labelled and unlabelled instances produced from the same set \citep{REALISTIC_EVAL_SEMI_SUPERVISED_2018, CD_UNLABELED_WU_2012} and diverse data sets with different quantities of labelled and unlabelled data~\citep{van2020survey} was already recommended to address this phenomenon. Analysing the impact of $p_\mathrm{u}$ on individual methods' performance is a way to adapt these findings to the needs of SSL evaluation under streaming scenarios.  \citet{le2019semi} presented a summarised analysis considering different ratios (from 90\% to 99\%) of unlabelled instances for streaming evaluation.

%\todo[inline]{This is where a reduced data stream definition will be}

Comparison of an SSL method against a fully supervised method based on removing some true labels is particularly challenging for the SSL method, as it makes the latter method rely on a lower number of labelled instances than the fully supervised method. Still, as shown in \citet{baron2015} and \citet{Masud2012}, an SSL stream mining method may provide accuracy comparable to or even competitive with a fully supervised technique under such circumstances. 
Further examples of works reporting that SSL approaches, even in such cases, can yield accuracy comparable to purely supervised learning are provided in the study \citet{REALISTIC_EVAL_SEMI_SUPERVISED_2018}, which is focused on the evaluation of deep SSL methods in a batch setting.

It is important to observe that $F_\mathrm{U}(\Psi,p_\mathrm{u})$ can be created from an initially available data stream, which could be either fully or partially labelled data stream. While we propose that an SSL method executed on $F_\mathrm{U}(F_\mathrm{L}(\Psi[T_\mathrm{min},T_\mathrm{max}]),p_\mathrm{u})$ is compared with a fully supervised method $F_\mathrm{L}(\Psi[T_\mathrm{min},T_\mathrm{max}])$, this does not exclude the use of a partially labelled original stream $\Psi$.
In particular, the SSL method can use both originally unlabelled instances and unlabelled instances caused by the use of $F_\mathrm{U}$.
%this approach is not always possible, as many real data streams are partly labelled only. 
Let us note that constraining SSL methods to make them use only those unlabelled instances which were originally labelled, would  not reflect the real needs and opportunities provided by SSL techniques.

% \subsubsection{The evaluation scenario based on removing unlabelled instances}
\subsubsection{Evaluation based on removing unlabelled instances}
Another way of comparing the performance of a fully supervised method with the performance of an SSL method is based on removing unlabelled instances. Unlike the former approach, under this scenario, the initial data stream has to be a partially labelled data stream $\Psi$, rather than fully labelled. The objective of the evaluation is to verify the merits of using unlabelled instances by comparing results attained on the partially labelled $\Psi$ with the results provided by a fully supervised method on a fully labelled stream $F_\mathrm{L}(\Psi)$ i.e. the $\Psi$ stream constrained to fully labelled instances. 
Indeed, the interest in semi-supervised learning is partly driven by the abundance of unlabelled data combined with scarce labelled data.
In such cases removing unlabelled instances is acceptable rather than  removing already scarce labels.
% It is crucial to notice that in such cases, it is this scenario i.e. the scenario of removing unlabelled instances that is feasible rather than the scenario of removing already scarce labels.

Among others, \citet{REALISTIC_EVAL_SEMI_SUPERVISED_2018} removed unlabelled instances to verify whether the performance obtained by training a model on $X_L \cup X_U$ (i.e. union of labelled data and unlabelled data) is better than the performance observed on labelled instances $X_L$ alone. \citet{REALISTIC_EVAL_SEMI_SUPERVISED_2018} observed that such a baseline is also frequently reported  in other studies. %By proposing the aforementioned comparison, we aim to extend this approach to delayed labelled data streams. 

The comparison of SSL methods exploiting both labelled and unlabelled parts of data streams to fully supervised methods which discard unlabelled data was performed in the study proposing a semi-supervised SVM learning framework \cite{ zhang2009mining}. The SSL methods proposed in the study outperformed the methods discarding unlabelled data. A related aspect of the impact the growing number of unlabelled training instances used by an SSL method on the overall accuracy was addressed in \cite{Masud2012}. The growth in the number of unlabelled training instances used by an SSL method resulted in accuracy improvements. 
%While the latter analysis did not focus on the comparison against fully-supervised method trained with labelled subset of examples only, \citep{Masud2012} illustrates the case of both comparing the SSL method applied to against the fully supervised method applied to the stream containing the same instances, yet fully labelled and 
In \cite{le2019semi}, the cluster-and-label method with pseudo-labeling was compared with its version without pseudo-labelling and found to outperform it for a number of synthetic and real data streams. This kind of comparison is one more example of investigating the benefits arising from including unlabelled training instances. Interestingly, the original data streams were fully labelled. Hence, \citet{le2019semi} illustrate the case of removing some labels from a fully labelled data stream first and considering or not unlabelled instances in pseudo-labelling next. 

\subsubsection{The benefits of both evaluation scenarios}

\begin{table}
\caption{Two scenarios of obtaining reference fully supervised baselines for the comparison of the performance  of SSL methods (SSM) with fully supervised methods (FSM)}
\begin{tabular}{|p{1.4cm}|p{1.1cm}|p{2.8cm}|p{5.7cm}|p{1cm}|}
\hline 
Scenario & Original method & Modified stream and method & Key features & Sample works \\ 
\hline 
\makecell[l]{Label\\ removal} & FSM& $F_\mathrm{U}(\Psi[T_\mathrm{min},T_\mathrm{max}],p_\mathrm{u})$ SSM & \begin{itemize}
    \item suitable when the source data stream is fully labelled 
    \item the number of labelled instances available for SSL methods lower than for fully supervised setting
    \item the same number of instances available for both fully supervised and SSL methods 
    \item fully supervised setting provides an expected upper bound for SSL model performance
\end{itemize}
& \cite{baron2015,dyer2014,van2020survey,le2019semi,Masud2012, CD_UNLABELED_WU_2012} \\ 
\hline 
Unlabelled instance removal & SSM & $F_\mathrm{L}(\Psi[T_\mathrm{min},T_\mathrm{max}])$  FSM & \begin{itemize}
    \item suitable for evaluation made for real partially labelled data streams, for which complete labelling is not feasible
    \item the same number of labelled instances available for both fully supervised and SSL methods 
    \item fully supervised setting provides an expected lower bound for SSL model performance
\end{itemize}
& \cite{le2019semi,REALISTIC_EVAL_SEMI_SUPERVISED_2018, zhang2009mining} \\ 
\hline 
\end{tabular}     \label{tab:evaluation_scenarios}
\end{table}

%\todo[inline]{Replace features with formulas}

It follows from the related works that by a ``fully supervised baseline'' two different baselines are meant in different studies. The first of them, which we refer to as the {\em label removal scenario} relies on removing some labels from an initially fully labelled data stream. The other approach i.e. the {\em unlabelled instance removal} scenario is to use a  partially labelled data stream to remove unlabelled instances from it. Both scenarios can be used to develop a fully labelled data stream and use it for supervised learning. 
% These two scenarios of developing data for the comparison of supervised vs. SSL methods are summarised in Table \ref{tab:evaluation_scenarios}. 
Using two different approaches can  in fact be justified by the merits of each of them, which we summarise in Table~\ref{tab:evaluation_scenarios}. 

In the {\em label removal} scenario, the fully supervised approach is assumed to yield an upper bound for the SSL performance, as the SSL model relies on a lower number of labelled instances. Moreover, the SSL model can access the same number of instances as its fully supervised counterpart, but true labels of some of these instances are not available for the SSL method. Hence, the benefits of using unlabelled instances are not expected to surpass the benefits of using the same instances provided with true labels. On the other hand, the {\em unlabelled instance removal scenario} means that all the labelled instances are available for both the fully supervised and SSL methods. %  fully supervised methods and SSL methods.
% \todo[inline]{Heitor: I removed the mention to active learning in here. My impression is that using active learning in a pure SSL environment is not possible, unless we are considering something like the fully supervised methods have access to a randomly selected number of labelled instances, while the SSL algorithm can actively choose them.}
% \todo[inline]{@Heitor: I am fine with it. Indeed, having active learning here could be too confusing. However, I think we may leave active learning in the evaluation algorithm. Or not, because we should then consider also other alternative approaches in the evaluation algorithm we propose? What do you think?
% @Maciej: that is a good question, let's leave it in the evaluation algorithm}
In addition, an SSL method may benefit from unlabelled instances available for this method only. This means that in the case of the unlabelled instance removal scenario, we intuitively expect the performance of fully supervised models to be worse than the performance of SSL models. 
Let us observe that  in the case of both expected upper and lower bound, the performance of the supervised method can be in some cases unexpectedly, respectively, worse and better  than the performance of an SSL method. These dependencies are not rigid due to multiple factors, some of which are: a) possible superiority of the supervised part of an SSL method over the fully supervised method, b) noisy true labels disturbing fully supervised learning, c) the use of data augmentation techniques %or active learning approaches
by some of the SSL methods increasing their chances of surpassing supervised methods despite the lower number of labelled instances.

To sum up, we propose both scenarios described above to be applied simultaneously to benchmark SSL methods against fully supervised methods. This is because comparison under a lower number of labelled instances and the same number of labelled instances accompanied by unlabelled instances are inherently different, and each provides additional insight into the functioning of SSL methods.

\subsection{Reference data streams}
\subsubsection{The selection of data used for comparative analysis}
The evaluation of individual stream mining methods under consideration should be made on a benchmark set of data streams. Similarly to other stream mining studies, we propose that evaluation performed with real data streams should be accompanied by  evaluation performed with synthetic data streams including the streams for which predefined concept drift events, including the periods affected by gradual concept drift, can be defined. 
% It is a common practice to use both real and synthetic data streams for the evaluation of stream mining methods. In particular, t
The evaluation of both synthetic and real data streams is a common practice in works proposing new stream mining methods~\citep{zhang2009mining,LEVERAGING_BAGGING_2010,ARF,SRP,zhang2009mining}. 
% Evaluation procedures such as continuous re-evaluation of models under delayed labelling settings \citep{grzenda2019delayed}, and surveys in the field of stream mining such as the survey on data preprocessing for stream mining \citep{DS_PREPROCESSING_2017}.

By definition, the evaluation requires multiple partly labelled delayed data streams to be included. However, synthetic data streams are typically fully labelled and rely on immediately available labels. This includes synthetic data streams frequently used in the evaluation of stream mining methods such as Agrawal \cite{grzenda2019delayed}, Hyperplane \cite{grzenda2019delayed}, LED \cite{grzenda2019delayed, le2019semi}, and Random Tree \cite{le2019semi}. 
As proposed in \citet{le2019semi}, labels from their instances can be removed with probability $p$ to provide a partly labelled data stream. 
% For every data stream under consideration, this provides what we call in this study a reduced partially labelled data stream $F_\mathrm{U}(\Psi[T_\mathrm{min},T_\mathrm{max}],p_\mathrm{u})$.

In their recent study, \citet{le2019semi} proposed establishing a baseline to evaluate semi-supervised learning methods in data streams. Importantly, this includes the extension of the MOA framework, which enables such evaluation. Even though this proposal does not consider the delayed labelling, but immediate labelling only, it can serve as a starting point for defining a baseline set of delayed data streams and developing  software serving evaluation needs. Data streams for which no natural delay exists, including all the synthetic data streams listed above, can be converted to delayed ones by adding a fixed delay~\citep{grzenda2019delayed}. 

To sum up, some reference data streams can be developed under the label removal scenario from their fully labelled versions, but also from real data streams. In this way, partially labelled data streams can be developed. Next, fully labelled data streams can be developed by applying the unlabelled instance removal scenario to the former streams. As a consequence, the results of such studies can be compared to the studies already made under a fully labelled setting for the original data streams.
Such data streams ideally should be accompanied by real partially labelled delayed data streams illustrating the abundance of unlabelled data.
% \todo[inline]{@Heitor: I have added the paragraph above to have some recommendation here and link the previous sections at the same time. I am looking forward to your opinion on it. 
% @Maciej: I agree with this, it provides a good summary on how data streams can be}

%Therefore, let us propose that a baseline benchmark includes the data streams listed in table 
%\todo[inline]{Shall we define a set of delayed data streams here? Actually we could combine the ideas from \citet{le2019semi} and \citet{grzenda2019delayed}.}

\subsubsection{Key aspects of the evaluation process}
Let us observe that for the evaluation of SSL methods to be fair, it is important to document all the assumptions and limitations it relies on, but also alternative approaches. Let us first discuss some of the assumptions which may have a potentially significant impact on the interpretation of evaluation results and on the evaluation process needed.

First of all, in some studies, an assumption can be made that the number and distribution of classes in the labelled and unlabelled parts of a data stream are the same. In some tasks, such as binary classification, in which the probability that an instance has no label depends neither on the instance data nor the true label, this approach can be justified.
However, as pointed out in \citet{REALISTIC_EVAL_SEMI_SUPERVISED_2018}, the predictive performance of SSL techniques can degrade drastically when the assumption of equal distribution of classes in the labelled and unlabelled parts of a data set is not met.

Another important aspect of the evaluation is whether hyperparameter tuning aimed at finding the best settings of the individual stream mining methods under comparison has been performed or not. In the case of stream mining methods, unlike in the case of batch methods such as the methods analysed in \citet{REALISTIC_EVAL_SEMI_SUPERVISED_2018}, a single pass over the data is expected i.e. every instance should be inspected at most once \citep{BOOK_MLDS_2017}. Hence, in contrast to batch learning, hyperparameter tuning should not rely on multiple runs with the same data. 
%Whether this approach was actually applied or not we suggest should be one of the assumptions explicitly documented in the performance analysis of SSL methods applied to data streams. 
Whether such multiple runs were applied or not, we suggest it should be one of the solutions explicitly documented in the performance analysis of SSL methods applied to data streams.
Not surprisingly, comparing a method for which large-scale hyperparameter tuning was carried out before its ultimate performance assessment to methods not tuned in such a way may be misleading. 
% On the other hand, hyperparameter tuning such as tuning the size of an ensemble can be an integral part of the main stream mining method. This kind of hyperparameter tuning  %to fully respect the assumptions of stream mining. 
% can fully respect the assumptions of stream mining including the use of one pass over the data.

One more important aspect of the evaluation is sensitivity analysis i.e. the analysis of the impact of varied hyperparameter settings on the performance of the methods under comparison, which is frequently illustrated with the plots showing the impact of parameter settings on the accuracy of the models. The sensitivity analysis of this kind was performed in \cite{dyer2014} and \cite{Masud2012}. When the method relies on fixed parameter settings not evolved during stream processing, such sensitivity analysis provides an insight into the resilience of the methods to varied, including possibly suboptimal parameter settings.

Furthermore, as noted by \citet{REALISTIC_EVAL_SEMI_SUPERVISED_2018}, a rarely reported baseline against which to compare SSL approaches is transfer learning. More precisely, the authors suggested comparing results attained with the SSL method to the results observed when a model was trained with a different (but similar and large) data set to be tuned with the small data set representing the ultimate task in the next stage. Not surprisingly, whether this approach can be applied, depends on the availability of similar data making the training of the initial model possible. 

% In the simplest case of two data streams with the same number and distribution of a fully supervised stream mining model developed with fully labelled data stream of a similar data stream can be adapted next to the \todo[inline]{add notation here}.
In a stream setting, a similar data stream may be a data stream containing instances collected from a similar sensor. Hence, to fully document possible alternatives, the availability of related data streams which could be used to provide initial stream mining models in a transfer learning setting could be checked. Nevertheless, transfer learning is a challenging approach for online scenarios~\citep{zhao2014online}.
%as discussed in Section \ref{sec:discussion}. 

\subsubsection{Key features of data streams}

\textbf{The impact of label latency.}
%   \item Which prediction to use when evaluating correct/incorrect predictions? First? Last? \textbf{Special case:} \textbf{Continuous re-evaluation}~\cite{grzenda2019delayed}
%     \todo[inline]{Maciej: I think we could say that we should evaluate these  predictions, which are applicable for a problem i.e. depending on the problem we could rely on: first predictions, final predictions (not very real-life approach, but reasonable to refer to a traditional evaluation) or intermediate performance measures. In the case of intermediate measures we could refer to the discussion from our IJCNN 2020 paper and what we proposed there. }
    Much of the research on supervised learning for delayed data streams had focused on evaluating predictions made for the instances when they were received from a data stream~$\Psi$ i.e. initial predictions. In contrast to the works considering label latency, immediate labelling studies assume that an instance's true label is available immediately after this instance. In such cases, the  test-then-train approach is frequently applied.
    %, in which prediction made for an instance immediately precedes the arrival of a true label. 
    This approach, when adopted to delayed labelling, suggests evaluating  final predictions i.e. predictions made for the instances immediately before the arrival of the true labels of these instances. 
    %The distinction between these two categories of predictions is justified by the fact that a model used to make a prediction at the time of receiving an instance may change before the arrival of its true label and yield different predicted values in turn. In fact, models are expected to evolve in response to the growing availability of the data and concept drift.
    
    % \todo[inline]{Heitor: The connection between this paragraph and the previous discussion on continuous re-evaluation could be explicit. For now, I just cited again the dami and ijcnn papers.
    % @Heitor: I think the text looks OK now. Please check again after my changes. I must say I am a bit lost in changes w have applied here :-)
    % @Maciej: Should be alright now, you are not lost alone! }
    Considering the needs of the evaluation of SSL methods for delayed data streams, let us observe that it should be focused on initial predictions.
    However, final predictions should also be included to reveal to what extent models evolve and predictions change in turn whilst waiting for true labels~\citep{grzenda2019delayed,grzenda2020ijcnn}.
    % Furthermore, as discussed in \citep{grzenda2019delayed,grzenda2020ijcnn}, for every problem for which periodic predictions can be exploited in practice, intermediate performance measures capturing the performance of both initial predictions and their updates produced by a model whilst waiting for a true label should be reported along with initial predictions. In this way, both the ability of the method under consideration to produce high quality first time predictions and their possible updates will be verified.

\textbf{Class label distribution}.
One more factor that may have a substantial impact on the performance of SSL methods is the presence of instances related to new classes (concept evolution~\cite{BOOK_MLDS_2017}) in the unlabelled data. \citet{REALISTIC_EVAL_SEMI_SUPERVISED_2018} showed that in such cases performance can even be degraded compared to not using any unlabelled data at all. In the case of stream mining methods, the ability to deal with novel classes is frequently assumed. Still, whether and if so how many examples of different classes were present in the form of both labelled and unlabelled instances should be documented in the evaluation of individual methods, both fully supervised and semi-supervised.

\textbf{Volume of unlabelled data.}
% \todo[inline]{Heitor: I added this first sentence to clarify that we are not assuming that \textbf{only} more unlabelled data is sufficient. I also added an example citing S3VMs }
A good match between the data characteristics and the SSL method biases is required to allow improvement by leveraging unlabelled data. On top of that, SSL methods' merits depend on the volume of unlabelled data available to them. A larger volume of unlabelled data can raise the confidence of such methods, for example, those that rely on ensuring that the decision boundary must pass through low-density areas in the input space~\cite{bennett1999semi}. 
% The merits of SSL methods depend on the volume of unlabelled data available for these methods. 
Hence, for the evaluation of an SSL method to be complete, an analysis of the impact of the volume of unlabelled data on the performance of the method should be included. Not surprisingly, when the number of unlabelled instances available for the method is minimal, the performance advantage of SSL over fully supervised methods is unlikely to be significant, if any. Some of the works made in the immediate labelling setting %,  while not relying on delayed labels 
have already analysed the impact of a growing number of unlabelled instances on the performance of the methods they proposed~\citep{Masud2012}. 

\citet{REALISTIC_EVAL_SEMI_SUPERVISED_2018} showed that the sensitivity of different SSL approaches to the amount of labelled and unlabelled data significantly varies. This finding, while observed for batch learning techniques, indicates that accuracy gains arising from the use of unlabelled data must be interpreted with caution. In particular, the superiority of a SSL method following from  experiments not including an evaluation of the impact of the proportion of unlabelled data on the performance of this method should be considered to be true for the quantity of labelled and unlabelled data used in the experiment, rather than for other ratios of unlabelled data.

% Let us observe that an analysis of t
The influence of the number of unlabelled instances can be observed for both real and synthetic streams. Even though the number of unlabelled instances for individual concepts may be strictly constrained by the data,
% real process providing a data stream
 it is possible to use only a subset of the available unlabelled instances in some of the runs of the method under consideration. Hence, individual method runs may rely on a different number of unlabelled instances and possibly different proportions of labelled vs. unlabelled instances. 
% In this way, the impact of both the number and the proportion of unlabelled instances on the performance of a SSL method can be determined. 

\subsubsection{Additional evaluation of active learning methods}
%\todo[inline]{I think, we can discuss this topic here i.e. after the basic data selection and performance measures are discussed}
Some SSL methods rely on active learning (AL). Active learning can be used not only to increase the availability of labelled data, but also to contribute to model adaptation to concept drift. A method relying on active learning to obtain additional labelled instances when concept drift is detected was proposed {\em inter alia} in \cite{REIS:2016}. When active learning becomes a part of an SSL method, additional evaluation of the method has to be considered. It is important to note that  comparison of active learning models vs. models trained on initially available labelled data  should only take into account the cost of obtaining additional labels by the active learning method.  In particular, in the case of the active learning method, the superiority of the method in terms of the performance of its predictions is not sufficient to confirm the actual improvement offered by the method over a method not requesting extra labels from an oracle such as a human expert.

The evaluation of the cost of obtaining extra labels from an oracle can be made a) in an on-line manner to control the number of requests for additional labels, in order not to jeopardise the benefits of the SSL/AL method and b) in an off-line manner i.e., calculated after the active learning method has been executed. When active learning methods are considered, which is problem-dependent, the problem of selecting the best method for a data stream or  set of data streams can be defined as a multi-objective optimisation problem, as both the performance of the method and the cost of obtaining extra labels have to be considered.

Recent studies on the evaluation of stream mining methods for delayed data streams \citep{grzenda2019delayed,grzenda2020ijcnn} reported that the accuracy of initial predictions is typically lower than the accuracy of final predictions. This phenomenon was observed both for synthetic and real data streams. 
This difference between the accuracy of initial and final predictions was even more significant for concept drifting data streams. This is because a more recent model benefits from a larger number of labelled instances, possibly reflecting recent changes in the underlying process \citep{grzenda2019delayed}. Therefore, for the evaluation of active learning approaches to be realistic, not only the cost of obtaining additional labels from an oracle but also the latency with which these labels are available should be considered. This latency cannot be neglected, especially when a human expert is assumed to be involved in the labelling process.
If this latency of obtaining additional labels were neglected, the evolution of a model benefiting from these labels would be assumed to be faster than actually possible. As a consequence, taking into account the results of the aforementioned studies, the value of performance measures reported for active learning while not considering labelling latency could potentially be unrealistically superior to the measure values reported for other methods.

To sum up, when a SSL method relies on the active learning paradigm, a recommendation can be made to both report the cost of obtaining extra labels and consider in the evaluation of the method the latency of obtaining additional labels from the method.

%\todo[inline]{We could define here a weighted function to take into account both the performance measure value and the cost of obtaining extra labels in active learning approach. An active learning approach yielding higher performance e.g. accuracy is not necessarily better, as the cost of extra labels may be not justified. @Heitor: 1) Shall we try to formally define the problem? 2) Shall we look for the formal approaches to comparing active vs. supervised already considering the cost of extra labels}

%\subsection{Comparison to transfer learning approaches}

% At 

% \todo[inline]{Mention that at least similar data streams could be sought to be applied with transfer learning}

% \todo[inline]{Consider class distribution problem mentioned by Oliver et al.}

\subsection{Unified fair evaluation}

Taking into account all the aforementioned aspects of the evaluation of SSL methods applied to delayed partially labelled data streams, let us propose  Alg.~\ref{alg:main_evaluation}  for such evaluation.
Importantly, the algorithm aims to show logical data flow rather than its physical implementation. Similarly to the seminal work on Hoeffding trees \cite{VFDT:2000}, additional measures can be applied at the implementation stage to reduce the computational load and storage needs of Alg.~\ref{alg:main_evaluation}, some of which are outlined below. 
The input for the algorithm is the set of reference data streams, which are expected to include both real and synthetic data streams. The algorithm starts by determining dependent data streams. As discussed above, any evaluation is constrained to a certain time period and the set of instances and labels from this period.

In Alg.~\ref{alg:main_evaluation}, two categories of $\Psi_q$ streams can be used i.e. fully labelled data stream as used in the label removal scenario, or a partially labelled data stream, used as an input in the unlabelled instance removal scenario. We suggest that both cases can be unified i.e. in both cases, the evaluation can include the input data stream and its labelled part only. Moreover, a particular disadvantage of comparing the performance of an SSL method observed on partially labelled $\Psi_s$ with the performance of a fully supervised method applied to $F_\mathrm{L}(\Psi_\mathrm{s})$ is the fact that the performance of both methods is analysed for only one proportion of labelled and unlabelled instances - already present in the input $\Psi_s$ stream. Hence, we propose testing the impact of removing some of the labels on both methods, % also in 
including the unlabelled instance removal scenario. As a consequence, for every $\Psi_q$ data stream, two categories of fully labelled data streams i.e. $\Psi_\mathrm{UFS}$ and $\Psi_\mathrm{LFS}$, providing an expected upper bound and lower bound respectively for the SSL method are developed. In the case of $\Psi_\mathrm{LFS}$ streams, the number of such streams matches the number of different $p_\mathrm{u}$ settings controlling the number of removed true instances. The partially labelled data streams are used to evaluate SSL methods $M_\mathrm{SSL}$, while the fully labelled data streams are used to evaluate $M_\mathrm{FS}$ methods. By a method a combination of stream mining method and its hyperparameter settings is meant. In this way, sensitivity analysis of individual methods can be performed. Ideally, both real and synthetic data streams including the streams with known presence of concept drift should be represented in the reference stream sections.

As far as the main instance loop operating on the instances of a single stream is concerned, let us emphasise that we take into account the initial predictions i.e. predictions made at the time of receiving instance data, final predictions i.e. predictions made at the time of receiving a true label, and periodic predictions. Furthermore, if  active learning methods are included in the evaluation, additional labels received on request from an oracle can be included and possibly used to update a model at the time of actually having them available. Last, but not least, we propose  these labels to be used for updating a model, but not for updating performance indicators, as the performance indicators should  rely entirely on the performance observed on input instances. This is because the distribution of instances for which additional labels are requested is not likely to match the distribution of $\mathbf{x}$ examples. Furthermore, by performance indicators both the indicators aggregating the similarity of predicted and true labels, including accuracy, $\kappa$, $\kappa^{+}$, and intermediate performance measures~\cite{grzenda2020ijcnn} and indicators revealing resource consumption, such as computation time and memory use are meant. In the former case, the assessment of initial, periodic and final prediction may reveal varied abilities of individual methods to evolve the models before true label arrival. In parallel, stream statistics including label latency histograms, distribution of classes, and the volume of labelled and unlabelled data can be collected.

As far as implementation aspects are concerned, all the streams present in $\Omega_q$ sets can be developed in parallel. In particular, multiple runs for every $p_\mathrm{u}$ and stream mining method combination can also be executed in parallel. Furthermore, all dependent streams can be gradually produced and processed in parallel without the need to store all their instances and labels. In contrast, every time a new instance or a label of an instance arrives, it  may or may not, depending on whether it is included in a dependent stream, become a part  of the dependent stream and be processed in the instance-based loop of the algorithm for this dependent stream. 

\begin{algorithm}
\caption{Evaluation of semi-supervised   methods under delayed labelling setting
\label{alg:main_evaluation}}
\small
\SetNoFillComment
\KwIn{
$\{\Psi_q[T_\mathrm{min}(q),T_\mathrm{max}(q)]: q=1,\ldots,Q\}$ - reference stream sections, $\{p_i: i=1,\ldots,P\}$ - label removal probabilities, $R$ - the number of runs, $M_\mathrm{SSL}$ and $M_\mathrm{FS}$ - semi-supervised and fully supervised stream mining methods to evaluate, respectively
}
\KwData{
$\mathcal{S}_1,\mathcal{S}_2,...$ - data stream;
$L$ - list of examples ($\{(\mathbf{x}_k,?)\},\ldots$), containing the data of instances awaiting their true labels, 
%and $l$ is the number of other labelled instanced received since $t(\{(\mathbf{x}_k,?)\})$; 
$P(k)$ - list of timestamped predictions made for $\mathcal{S}_i=\{(\mathbf{x}_k,?)\}$,  
%$\Lambda(t,b)$ - performance measure such as accuracy observed by time t for predictions available in $b$-th bin, 
%the list contains tuples $(\hat{\mathit{y}},t,b)$, where $\hat{\mathit{y}}$ denotes prediction, $t$ - the time of generating it and $b$ - the bin it belongs to, respectively, while $b=-1$ denotes a placeholder bin i.e. a bin containing periodical predictions until true label $\mathit{y}_{k}$ arrives i.e. until these predictions can be shuffled into proper bins $b=1,\ldots,B$ associated with subperiods between $t(\mathcal{S}_i)$ and $\tilde{t}(\mathcal{S}_a)$ ; 
%$E(b), b=0,\ldots,B+1$ - evaluators for individual bins calculating performance measures, e.g. accuracy, 
$h_i$ - the prediction model after processing $i$ instances,
$M(\Psi)$ - applicable methods i.e. $M_\mathrm{SSL}$ and $M_\mathrm{FS}$ for partly and fully labelled $\Psi$, respectively
%trained with labelled instances $\mathcal{S}_j; j=1,\ldots,i-1$
%, $C(i),C_\mathrm{L}(i)$ - the number of calculated predictions following initial prediction, after processing $i$ instances for all and labelled instances, respectively
}

\Begin{
% \tcc{For every stream in benchmark set}
\For{$q=1,\ldots,Q$}
{
$\Psi_\mathrm{s}=\Psi_q[T_\mathrm{min}(q),T_\mathrm{max}(q)]$;
$\Psi_\mathrm{UFS}=F_\mathrm{L}(\Psi_\mathrm{s})$\;
$\Omega_q=\{\Psi_\mathrm{s},\Psi_\mathrm{UFS}\}$\;
%$C(0)=0$; $C_\mathrm{L}(i)=0$\;
\For{$p_\mathrm{u} \in \{p_i: i=1,\ldots,P\}$}
{
\For{r=1,\ldots,R}
{
$\Psi_\mathrm{SSL}=F_\mathrm{U}(\Psi_\mathrm{UFS},p_\mathrm{u})$; $\Psi_\mathrm{LFS}=F_\mathrm{L}(\Psi_\mathrm{SSL})$\;
$\Omega_q=\Omega_q\cup  \{\Psi_\mathrm{SSL},\Psi_\mathrm{LFS}\}$\;
}
}
\For{$\Psi \in \Omega_q$}
{
\For{$M \in M(\Psi)$}
{
$h_1=\phi$; $L=\phi$; $P=\phi$\;
\tcc{Instance loop}
\For{$i=1,\ldots$}
{
$\mathcal{S}_i$=fetchNext($\Psi$)\;
\tcc{New unlabelled instance arrived}
 \uIf{$\mathcal{S}_i= \{(\mathbf{x}_k,?)\}$}{
 %\uIf{$\neg L$.contains($k$)}{
  $L$.add($\{(\mathbf{x}_k,?)\}$)\;
  % }
  \tcc{obtain first time prediction}
  P($k$).addFirst($h_i(\mathbf{x}_k),t(\mathcal{S}_i)$)\;
   $h_{i+1}$=trainSSL($M,h_{i},\{(\mathbf{x}_{k},?)\}$\;
 }
 \Else{
 \tcc{$\mathcal{S}_i= \{(\mathbf{x}_k,\mathit{y}_k)\}$, i.e. a true label arrived}
 \tcc{if the label is a delayed label}
 \If{c($\mathcal{S}_i)=\mathrm{DELAYED}$}
 {

  \tcc{obtain final prediction}
  P($k$).addFinal($h_i(\mathbf{x}_k),t(\mathcal{S}_i)$)\;
 % $T=t(\mathcal{S}_i)$\;
%   \tcc{Update performance measures for every bin}
%   \For{$b=0,\ldots,B+1$}
%   {
%   $\Lambda(T,b)$=updPerformance($P(k),\mathit{y}_k,t(\mathcal{S}_i)$)\;
%   }
%   
\tcc{Calculate performance measures and stream statistics}
updPerformance($P(k),\mathit{y}_k,t(\mathcal{S}_i)$);   $L$.remove($k$)\;

%   \For{$m=1,\ldots, M$}
%   {
%   $\mathbf{v}=[\Lambda(T,0),\ldots,\Lambda(T,B)]$\;
 % $\Psi(T)_\alpha=\Psi_\alpha(\Lambda(T,0),\ldots,\Lambda(T,B))$\;
%   }

  \tcc{generate new periodic predictions for instances awaiting true labels}
  $L$.generateNewPredictions($h_{i}$)\;
 }
 
  \tcc{Update the model with SSL method for SSL streams or FSM method for UFS and LFS streams, based on delayed and active learning labels}
  $h_{i+1}$=train($M,h_{i},\{(\mathbf{x}_{k},\mathit{y}_{k})\}$)\;
  
%  }
%  \Else
%  {
%  \tcc{A label requested in active learning model arrived}
  
%     $h_{i+1}$=train($h_{i},\{(\mathbf{x}_{k},\mathit{y}_{k})\}$)\;
%   }
}
}
}
}
}
}
\end{algorithm}

%\todo[inline]{Add references to hyperparameter tuning in streaming setting}

% \subsection{Summary of differences compared to batch learning}
% \todo[inline]{Many aspects seem to be similar. So might be the survey. Hence, it may be reasonable to highlight the key aspects unique for the evaluation of stream mining methods. These are as follows: different notions of fully supervised baseline, no need to vary the volume of labelled data as we gradualyy process a data stream after all, no hyperparameter tuning,....
% + Emphasise the need for the evaluation to be realistic (referring to Oliver)}

% \input{6_tools.tex}

%\input{6_discussion_future.tex}
% \input{6b_perspectives.tex}

\section{CONCLUSION AND PERSPECTIVES} \label{sec:conclusion}
In this paper, we discussed semi-supervised learning from the perspective of delayed partially labelled data streams. This setting is a realistic representation of several real-world problems involving the application of machine learning for data streams. 
We present several aspects of SSL in this context, including: related problems; learning guarantees; classic batch methods and their online counter parts; and fair evaluation of SSL methods. 

% \section{Perspectives} \label{sec:perspective}

SSL methods have been explored (and continue to be explored) with varying levels of success in the batch setting, therefore it is worth reflecting upon how this success can be reflected in the settings of partially-labelled data streams. An explosive area to look at is that of deep learning where the interest in SSL techniques has increased rapidly in the community~\citep{van2020survey}. SSL is particularly relevant to these models which are notoriously data hungry, and often not enough labeled data is available. Data augmentation can also be useful in this setting: augmenting the training data with new examples artificially created from existing ones. A promising, albeit challenging, venue of research is the application of such techniques to unlabelled examples; some have already been proposed \cite{xie2019unsupervised}. Extending such methods to data streams poses several challenges, namely that of computational complexity. Augmentation strategies defined for batch data already require costly operations, such as SMOTE~\citep{SMOTE_2002} which was originally used to address class imbalance problems, but can also be used to augment the training data. Recently, \citet{CSMOTE} proposed a meta-strategy named continuous-SMOTE that perform the oversampling step only on a recent subset of instances. 

The application of deep neural networks to streaming is controversial. Although gradient descent is naturally instance-incremental (or, more commonly, minibatch-incremental), a large amount of labeled data is required, and often the convergence of gradient descent algorithms require many passes over the data, which is not possible a streaming setting. There are indeed approaches to apply deep learning models to streams~\citep{ashfahani2019autonomous,pratama2019weakly}, however the advancements in the field are lagging behind in comparison to the batch community. 

Transfer learning is an increasingly popular and powerful technique to improve the performance of learning one concept, given that an earlier concept has already been learned. There is an obvious connection here to learning in a concept-drifting data stream, but also more generally when trying to stream one concept making use of another existing already-learned task. \citet{zhao2014online} introduced a framework for online transfer learning, including algorithms to tackle domains of common and different feature spaces, as well as an algorithm to address concept-drifting streams. More recently, \citet{wu2017online} explored multiple homogeneous and heterogeneous sources for online transfer learning, however concept drift was not taken into account for the scenarios studied. We remark upon this topic because SSL can be leveraged for the application of transfer learning, since more accurate models can be produced even when labelled data is scarce in the target domain. In Section \ref{sec:comparative_analysis}, we mentioned that comparing the predictive performance of SSL methods against fully supervised methods that benefit from transfer learning leads to a realistic benchmark. However, this practice is yet to become popular as transfer learning is not widely used in streaming in comparison to batch settings. 

% As pointed out in Section \ref{sec:comparative_analysis}, comparing the predictive performance of SSL methods against fully supervised methods that also benefit from transfer learning leads to a realistic benchmark. However, transfer learning can also be used to improve SSL methods, not only to avoid a cold-start of the models, but also to assist in the online training. For example, suppose a set of IoT devices each of which has an individual streaming model to accomplish a specific task. The models can improve one another, for example, by assuming a learning by disagreement strategy~\cite{zhou2010semi}. 

% \subsection{The role of privacy in SSL}

% \subsection{Regression tasks} 
%\textbf{SSL for other machine learning tasks}. Finally, SSL is often associated with classification problems. Most of the SSL assumptions are only valid for classification tasks; for example, the low-density assumption cannot be directly translated to a regression task. 

Also most SSL tasks are focussed on classification, several SSL techniques can be adapted to regression \citep{van2020survey}, such as self-training, co-training, and learning by disagreement. One possible pathway for future research is the application of such SSL techniques in regression problems. 
% One could also seek to improve clustering algorithms by training a supervised method on a small amount of labelled data. 

% In the near future, we foresee more interest in methods to address delayed partially labelled data streams, including more adaptations of classic batch algorithms and semi-supervised concept drift detection. 

\begin{acks}
Maciej Grzenda: The project was funded by \grantsponsor{}{POB Research Centre for Artificial Intelligence and Robotics of Warsaw University of Technology}{}  within the Excellence Initiative Program - Research University (ID-UB).
\end{acks}

\bibliographystyle{ACM-Reference-Format}
\bibliography{references}

\end{document}